%% file: main.tex
\documentclass[10pt,twocolumn,letterpaper]{article}

\usepackage[pagenumbers]{cvpr}

\definecolor{cvprblue}{rgb}{0.21,0.49,0.74}
\usepackage[pagebackref,breaklinks,colorlinks,allcolors=cvprblue]{hyperref}

\input{header}

\def\paperID{*****} % *** Enter the Paper ID here
\def\confName{CVPR}
\def\confYear{2026}

\title{{\name}: Mimic with Correspondence-Aware Cascade-Transformer\\for Category-Free 3D Pose Transfer}

\author{Zenghao Chai$^{1}$ ~~~~ Chen Tang$^{2}$ ~~~~ Yongkang Wong$^{1}$ ~~~~ Xulei Yang$^{3}$ ~~~~ Mohan Kankanhalli$^{1}$ \\
$^{1}$School of Computing, National University of Singapore \\ $^{2}$MMLab, The Chinese University of Hong Kong  ~~~~~
$^{3}$Institute for Infocomm Research, A$^*$STAR\\
\tt \faGithub~{{\href{https://mimicat3d.github.io/}{https://mimicat3d.github.io/}}}
}

\begin{document}
\maketitle

\begin{abstract}

3D pose transfer aims to transfer the pose-style of a source mesh to a target character while preserving both the target's geometry and the source's pose characteristic. Existing methods are largely restricted to characters with similar structures and fail to generalize to \textbf{category-free} settings ({\eg}, transferring a humanoid's pose to a quadruped). The key challenge lies in the structural and transformation diversity inherent in distinct character types, which often leads to mismatched regions and poor transfer quality.
To address these issues, we first construct a million-scale pose dataset across hundreds of distinct characters. We further propose {\name}, a cascade-transformer model designed for category-free 3D pose transfer. Instead of relying on strict one-to-one correspondence mappings, {\name} leverages semantic keypoint labels to learn a novel soft correspondence that enables flexible many-to-many matching across characters. 
The pose transfer is then formulated as a conditional generation process, in which the source transformations are first projected onto the target through soft correspondence matching and subsequently refined using shape-conditioned representations.
Extensive qualitative and quantitative experiments demonstrate that {\name} generalizes plausible poses across diverse character morphologies, surpassing prior approaches restricted to narrow-category transfer ({\eg}, humanoid-to-humanoid).

\end{abstract}

\input{1-intro}
\input{2-related-work}
\input{3-method}
\input{4-exp}
\input{5-abl}
\input{6-conc}

{
    \small
    \bibliographystyle{ieeenat_fullname}
    \bibliography{main}
}

% WARNING: do not forget to delete the supplementary pages from your submission 
\input{0-supp}

\end{document}

%% file: header.tex
\usepackage{times}
\usepackage{epsfig}
\usepackage{graphicx}
\usepackage{amsmath}
\usepackage{amssymb}
\usepackage{booktabs}
\usepackage{enumitem}
\usepackage{dsfont}
\usepackage{algorithm, algorithmic}
\algsetup{linenosize=\small}
\usepackage{tablefootnote}
\usepackage[flushleft]{threeparttable} % for tablenotes
\usepackage{balance}
\usepackage{xcolor}
\usepackage{color,colortbl}
\usepackage{overpic}
\usepackage{wrapfig}
\usepackage[normalem]{ulem}
\usepackage{amsmath,esvect}
\usepackage{orcidlink}

\usepackage[accsupp]{axessibility}

\usepackage{mathtools}
\usepackage{mathrsfs}

\usepackage{lipsum}
\usepackage{caption}

\definecolor{tabblue}{HTML}{00b4d8}

\input{insbox.tex}
\newcommand{\name}{MimiCAT}
\newcommand{\dataname}{PokeAnimDB}

\usepackage{tikzlings}
\usepackage{tikzducks}

\usepackage{wasysym}

\usepackage{multirow}
\usepackage{multicol}
\usepackage{soul}
\setul{0.5ex}{0.3ex}

\newcommand{\myparagraph}[1]{\vspace{2mm}\noindent\textbf{#1}}

\usepackage{pifont} 

\newcommand{\xmark}{\ding{55}}

\usepackage{xspace}

\makeatletter
\DeclareRobustCommand\onedot{\futurelet\@let@token\@onedot}
\def\@onedot{\ifx\@let@token.\else.\null\fi\xspace}

\def\eg{\emph{e.g}\onedot} 
\def\ie{\emph{i.e}\onedot} 
 
\def\etc{\emph{etc}\onedot} \def\vs{\emph{vs}\onedot}
\def\wrt{w.r.t\onedot} 
\def\aka{a.k.a\onedot}

\def\vs{\emph{v.s}\onedot}

\makeatother

\DeclareMathOperator*{\argmin}{arg\,min}

\newcommand{\nkptsrc}{K_1}
\newcommand{\nkpttgt}{K_2}

\newcommand*{\Scale}[2][4]{\scalebox{#1}{$#2$}}%

\usepackage{fontawesome}

%% file: insbox.tex
\catcode`\@ = 11
\newdimen\@InsertBoxMargin
\newcount\@numlines    % sum: m_1+...+m_n
\newcount\@linesleft   % counter used when reading lines of \ParShape
\def\ParShape{%
    \@numlines = 0
    \def\@parshapedata{ }% here we'll collect data for plain \parshape
    \afterassignment\@beginParShape
    \@linesleft
}%
\def\@beginParShape{%
    \ifnum \@linesleft = 0
      \let\@whatnext = \@endParShape
    \else
      \let\@whatnext = \@readnextline
    \fi
    \@whatnext
}%
\def\@endParShape{%
    \global\parshape = \@numlines \@parshapedata
}%
\def\@readnextline#1 #2 #3 {% #1 #2 #3 are: m_i, leftskip_i, rightskip_i
    \ifnum #1 > 0
      \bgroup  % I want to keep changes of \dimen0 and \count0 local
        \dimen0 = \hsize
        \advance \dimen0 by -#2  % \parshape requires left skip and
        \advance \dimen0 by -#3  % _length_of_line_ (not right skip!)
        \count0 = 0
        \loop
          \global\edef\@parshapedata{%
            \@parshapedata    % add to \@parshapedata:
            #2                % left skip
            \space            % a space
            \the\dimen0       % length of line
            \space            % another space
          }%
          \advance \count0 by 1
          \ifnum \count0 < #1
        \repeat
      \egroup
      \advance \@numlines by #1
    \fi
    \advance \@linesleft by -1
    \@beginParShape
}%
\newbox\@boxcontent     % box containing the picture to be inserted
\newcount\@numnormal    % number of leading lines to typeset normally
\newdimen\@framewidth   % width of the frame
\newdimen\@wherebottom  % position of frame's bottom
\newif\if@byframe       % true if we are just beside the frame
\@byframefalse
\def\InsertBoxC#1{%
  \leavevmode
  \vadjust{
    \vskip \@InsertBoxMargin
    \hbox to \hsize{\hss#1\hss}
    \vskip \@InsertBoxMargin
  }%
}%
\def\InsertBoxL#1#2{%
  \@numnormal = #1
  \setbox\@boxcontent = \hbox{#2}%
  \let\@side = 0
  \futurelet \@optionalparameter \@InsertBox
}
\def\InsertBoxR#1#2{%
  \@numnormal = #1
  \setbox\@boxcontent = \hbox{#2}%
  \let\@side = 1
  \futurelet \@optionalparameter \@InsertBox
}%
\def\@InsertBox{%
  \ifx \@optionalparameter [
    \let\@whatnext = \@@InsertBoxCorrection
  \else
    \let\@whatnext = \@@InsertBoxNoCorrection
  \fi
  \@whatnext
}%
\def\@@InsertBoxCorrection[#1]{%
  \ifx \@side 0
    \@@InsertBox{#1}{0}{{\the\@framewidth} 0cm}%
  \else
    \@@InsertBox{#1}{1}{0cm {\the\@framewidth}}%
  \fi
}%
\def\@@InsertBoxNoCorrection{%
  \@@InsertBoxCorrection[0]%
}%
\def\@@InsertBox#1#2#3{%
  \MoveBelowBox
  \@byframetrue
  \@wherebottom = \baselineskip
  \multiply \@wherebottom by \@numnormal
  \advance \@wherebottom by 2\@InsertBoxMargin
  \advance \@wherebottom by \ht\@boxcontent
  \advance \@wherebottom by \pagetotal
  \ifdim \pagetotal = 0cm
    \advance \@wherebottom by -\baselineskip  % ^ reduction
  \fi
  \advance \@wherebottom by #1\baselineskip
  \@framewidth = \wd\@boxcontent
  \advance \@framewidth by \@InsertBoxMargin
  \bgroup  % to keep changes of \dimen0 local
    \ifdim \pagetotal = 0cm
      \dimen0 = \vsize
    \else
      \dimen0 = \pagegoal
    \fi
    \ifdim \@wherebottom > \dimen0
      \immediate\write16{+--------------------------------------------------------------+}%
      \immediate\write16{| The box will not fit in the page. Please, re-edit your text. |}%
      \immediate\write16{+--------------------------------------------------------------+}%
      \vrule width \overfullrule
    \fi
  \egroup
  \prevgraf = 0
  \vbox to 0cm{%
    \dimen0 = \baselineskip
    \multiply \dimen0 by \@numnormal
    \advance \dimen0 by -\baselineskip
    \setbox0 = \hbox{y}%
    \vskip \dp0
    \vskip \dimen0
    \vskip \@InsertBoxMargin
    \ifnum #2 = 1
      \vtop{\noindent \hbox to \hsize{\hss \box\@boxcontent}}%
    \else
      \vtop{\noindent \box\@boxcontent}%
    \fi
    \vss
  }%
  \vglue -\parskip
  \vskip -\baselineskip
  \everypar = {%
    \ifdim \pagetotal < \@wherebottom
      \bgroup  % to keep some changes local
        \dimen0 = \@wherebottom
        \advance \dimen0 by -\pagetotal
        \divide \dimen0 by \baselineskip
        \count1 = \dimen0
        \advance \count1 by 1
        \advance \count1 by -\@numnormal
        \ifnum #2 = 1
          \ParShape = 3
                      {\the\@numnormal}   0cm   0cm
                      {\the\count1}       0cm   {\the\@framewidth}
                      1                   0cm   0cm
        \else
          \ParShape = 3
                      {\the\@numnormal}   0cm                  0cm
                      {\the\count1}       {\the\@framewidth}   0cm
                      1                   0cm                  0cm
        \fi
      \egroup
    \else
      \@restore@    % it's time to end everything
    \fi
  }%
  \def\par{%
      \endgraf
      \global\advance \@numnormal by -\prevgraf
      \ifnum \@numnormal < 0
        \global\@numnormal = 0
      \fi
      \prevgraf = 0
  }%
}%
\def\MoveBelowBox{%
  \par
  \if@byframe
    \global\advance \@wherebottom by -\pagetotal
    \ifdim \@wherebottom > 0cm
      \vskip \@wherebottom
    \fi
    \@restore@
  \fi
}%
\def\@restore@{%
    \global\@wherebottom = 0cm
    \global\@byframefalse
    \global\everypar = {}%
    \global\let \par = \endgraf
    \global\parshape = 1 0cm \hsize
}%
\ifx \documentclass \@Dont@Know@What@It@Is@
\else
  \let \pageno = \c@page
\fi

\catcode`\@ = 12

%% file: 1-intro.tex
\section{Introduction}

Recent advancements in computer graphics and vision have led to remarkable progress in modeling articulated 3D characters~\cite{leemove,song2025puppeteer,yun2025anymole,zhangmagicpose4d,raab2024single}. 
Among their key properties, articulated poses play a crucial role in conveying the behaviors and emotions of diverse types of characters.  
Once poses and animations are created, reusing them for novel characters and scenarios--{\aka} 3D pose transfer~\cite{yoo2024neural,wang2020neural,aigerman2022neural,muralikrishnan2025smf}--becomes highly desirable. 
The objective is to apply a reference pose from a source character to a target character while simultaneously preserving the unique characteristics of the target and the fidelity of the source pose.
However, transferring poses across different characters is highly non-trivial. 
Compared to humanoids, many characters possess distinct body structures and proportions, making reproducing similar behaviors challenging. 
The task is often dependent on experienced 3D artists, making it both expensive and slow.

\input{figure/teaser}

While early works~\cite{chen2022geometry,wang2020neural,wang2023zero,chen2021intrinsic} have shown promising results for pose transfer between characters with similar morphology ({\eg}, from a human to a robot) by learning keypoint- or vertex-level correspondences, they struggle to generalize across arbitrary character categories ({\eg}, transferring bird poses to humanoids).  
A key limitation is their reliance on one-to-one mappings to establish correspondences. Such mappings are often inadequate for modeling complex many-to-many relationships--for example, shall four limbs of a humanoid correspond to the two wings of a bird? This mismatch leads to significant transfer artifacts when dealing with characters of fundamentally different topologies.  
Moreover, existing approaches typically learn pose priors from human motion datasets~\cite{mahmood2019amass,mixamo2025,guo2022generating,wang2024towards}, making them prone to out-of-distribution failures ({\eg}, unnatural artifacts). For instance, transferring human poses to birds can yield degraded results due to the lack of knowledge about bird-specific motion patterns.  
These challenges arise because different characters exhibit highly diverse and complex structures ({\eg}, keypoints, skinning weights, and topologies), 
and their keypoints exhibit distinct rotation patterns conditioned on each character’s morphology and rigging design.
These factors collectively make cross-category 3D pose transfer a highly challenging problem.

To address the above challenges, we first construct a large-scale pose dataset containing $\sim$4.4 million pose samples across hundreds of diverse character categories (see Tab.~\ref{tab.datacmp}).
Using this dataset, we pretrain a shape-aware distribution model that captures the joint distribution of keypoint-wise rotations~\cite{yin2022fishermatch,sengupta2021hierarchical,mohlin2020probabilistic} for characters with varying skeletal structures. This pretrained distribution model serves as a regularization for pose transfer and prevents degenerate transformations.
Following this, we introduce {\name}, a transformer-based~\cite{vaswani2017attention} framework for category-free 3D pose transfer. {\name} features two cascaded transformer modules, which undergo two-stage training.
In the first stage, a \emph{correspondence transformer} learns a similarity matrix between two sets of keypoints with varying lengths.
Rather than relying on rigid one-to-one mappings, we incorporate semantic skeleton labels ({\ie}, keypoint names) to establish flexible many-to-many soft correspondences between structurally distinct characters. 
In the second stage, the estimated soft correspondences map initial transformations from the source to the target.
Conditioned on pretrained shape encoders~\cite{zhao2023michelangelo}, we formulate pose transfer as a conditional generation problem and employ a \emph{pose transfer transformer} to generate realistic target poses.
This stage employs a self-supervised cycle-consistency loss, eliminating the need for paired cross-category ground-truth.

To evaluate the transfer quality for category-free pose transfer, we establish a new benchmark based on cross-transfer cycle consistency.
% In addition, f
Following previous works~\cite{liao2022skeleton,wang2023zero,wang2020neural}, we also include humanoid-to-humanoid evaluations for comprehensive comparisons. 
Through qualitative and quantitative evaluation, we demonstrate that {\name} effectively transfer poses across 3D characters with significantly different morphologies, outperforming existing methods limited to structurallly similar characters.
% We conduct both qualitative and quantitative experiments to demonstrate that {\name} can effectively transfer poses across 3D characters with significantly different morphologies, outperforming existing methods that are restricted to structual-similar characters.
%
Once trained, our model seamlessly supports several downstream tasks, such as text-to-any-character motion generation.

In summary, our main contributions are as follows:
\begin{enumerate}
    \item We extend the 3D pose transfer task to a broader and more challenging setting, namely \emph{category-free pose transfer}, and construct a novel benchmark based on cycle-consistency evaluation to assess transfer quality.
    \item We build a large-scale dataset, {\dataname}, containing millions of character poses across diverse categories, which enables learning 3D pose transfer models under more practical and generalizable conditions.
    \item We design a novel cascade-transformer architecture, namely {\name}, that learns many-to-many soft correspondences between keypoints, enabling effective 3D pose transfer across characters with distinct structures.
    \item We achieve state-of-the-art pose transfer quality and demonstrate the strong potential of our framework for downstream applications. 
    % Code and dataset will be made publicly available for research purposes.
\end{enumerate}

\input{table/tab_stat_data}

%% file: figure/teaser.tex
\begin{figure}[t!]
    \centering
    \begin{overpic}
        [trim=0cm 0cm 0cm 0cm,clip,width=\linewidth,grid=false]{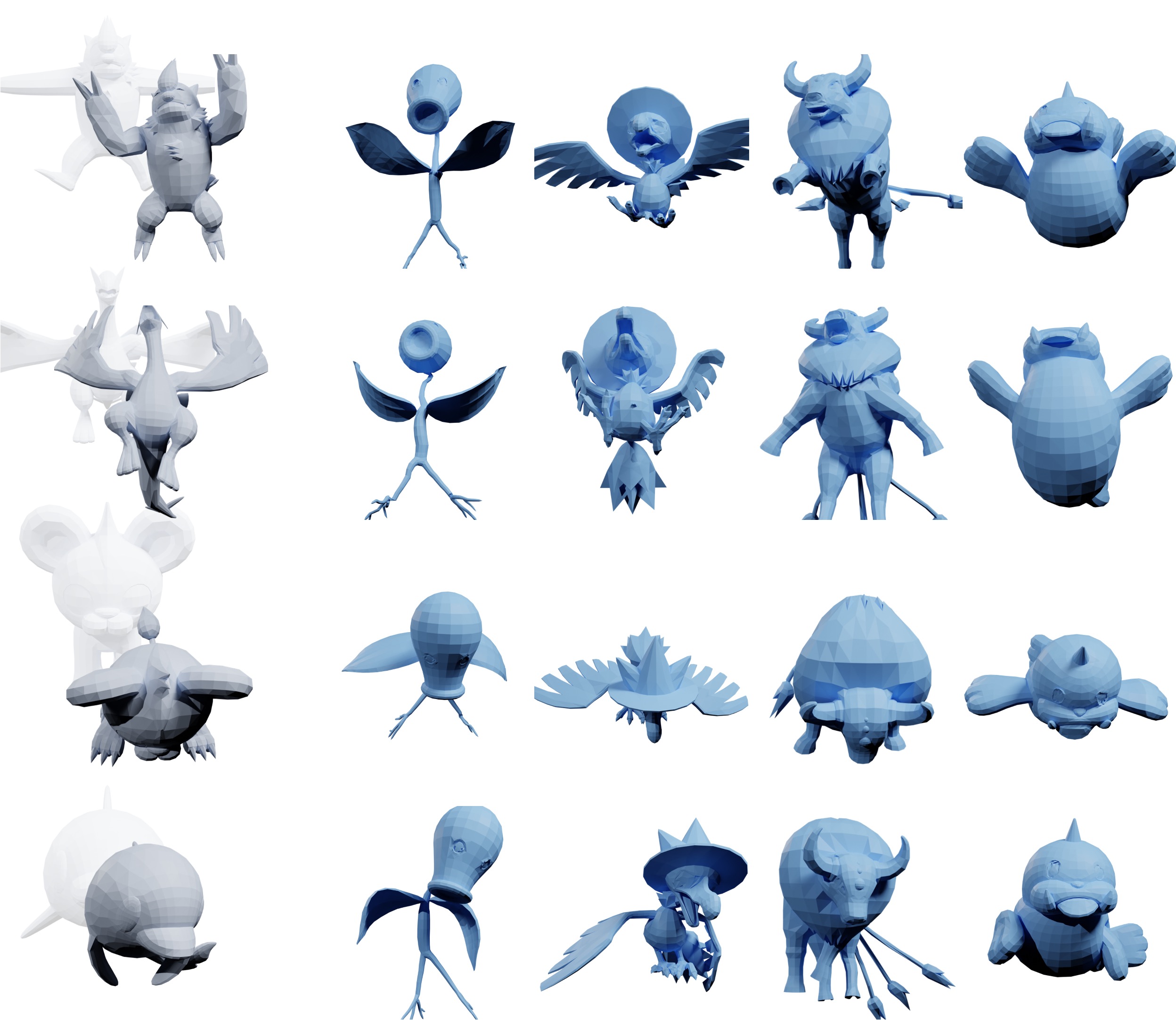}
        \put(5,84){\scriptsize \textbf{Source Pose}}
        \put(45,84){\scriptsize \textbf{Category-free 3D Pose Transfer}}
    \end{overpic}
    \vspace{-2em}
    \caption{\textbf{{\name} for category-free 3D pose transfer.}
    Given source character with desired poses (\textit{left}), our model faithfully transfers the given pose to the target characters (\textit{right}) across completely different categories, proportions and topologies, without requirement of manually labeled correspondence. 
    }
    \vspace{-10pt}
    \label{fig.teaser}
\end{figure}

%% file: table/tab_stat_data.tex
\begin{table}[t!]
\centering
\caption{\textbf{Statistics of character motion datasets.}
Our dataset is compared against representative publicly available motion datasets in terms of rigging, number of characters, and motion coverage.}
\vspace{-1em}
\setlength{\tabcolsep}{8pt} % column separation
\resizebox{\linewidth}{!}{%
\begin{tabular}{l|c|c|c|c}
\toprule
\textbf{Dataset} & \textbf{Rigging} &  \textbf{\#Characters} & \textbf{\#Motions} & \textbf{\#By Artist}\\ 
\midrule
Mixamo~\cite{mixamo2025}            & \checkmark                    & $118$              & $2{,}000$   &  \checkmark                          \\
AMASS~\cite{mahmood2019amass}             &  \checkmark                       & $346$ & $11{,}451$   & \checkmark    \\
TruebonesZoo~\cite{truebones2022}        &  \checkmark                & $70$                     & $1{,}000$   & \checkmark                           \\
PlanetZoo~\cite{wu2022casa}        &  \checkmark                & $251$                     & $251$    & \xmark                          \\
DT4D-A~\cite{li20214dcomplete} & \xmark         & $71$                     & $1{,}772$    & \checkmark      \\ 
\midrule
{\dataname} (\textbf{Ours})     & \checkmark & $\mathbf{975}$ & $\mathbf{28{,}809}$ & \checkmark \\ 
\bottomrule
\end{tabular}
}
\label{tab.datacmp}
    \vspace{-10pt}
\end{table}

%% file: 2-related-work.tex
\section{Related Work}

\myparagraph{Automatic 3D Object Rigging.}
3D Rigging, the task of predicting skeletons and skin weights for a 3D mesh, is a fundamental problem in computer graphics~\cite{zhang2025physrig,zhang2025one,mosella2022skinningnet,xu2020rignet,tagliasacchi2009curve}. Driven by the accessibility of open-source datasets, learning-based methods have outperformed traditional manual rigging. However, most public datasets either focus on static meshes~\cite{xu2020rignet,deng2025anymate,song2025magicarticulate,xu2019predicting}, or provide animations restricted to narrow categories such as humanoids~\cite{loper2015smpl,luo2023rabit,mixamo2025} or quadrupeds~\cite{li20214dcomplete,wu2022casa}.
Building upon these datasets, automatic rigging methods~\cite{xu2020rignet,mosella2022skinningnet,song2025magicarticulate,liu2025riganything,deng2025anymate,pan2021heterskinnet} aim to predict skeletons and skin weights directly from geometry. To enable animating previously static characters through predicted skeletons, such methods often remain dependent on handcrafted correspondences or manually curated motions.

\myparagraph{Correspondence learning} constitutes a fundamental challenge in vision and graphics~\cite{zhang2020cross,wang2020combinatorial,wang2019learning,du2025emergent,xu2022hierarchical,gat2025anytop}, and is particularly crucial for pose transfer. However, establishing reliable mappings between non-rigid meshes remains highly challenging. Dense correspondence methods~\cite{pai2021fast,magnet2022smooth,song20213d} predict per-vertex mappings by optimizing complex energies, but are computationally expensive, unstable, and generalize poorly beyond limited categories.
Recent works~\cite{liao2022skeleton,chen2023weakly,yoo2024neural,muralikrishnan2025smf,martinelli2024skeleton,aberman2020skeleton} instead consider sparse correspondences at the vertex or keypoint level, often using self-supervised learning to avoid the need for ground-truth dense annotations. These approaches are more flexible and can leverage existing rigging datasets~\cite{xu2020rignet,li20214dcomplete}, but typically assume one-to-one mappings and remain restricted to intra-categories, struggling to handle cross-category correspondences ({\eg}, between humanoids and quadrupeds).  

In contrast, we construct a dataset spanning diverse categories and character-level animations, and leverage semantic keypoint annotations to supervise many-to-many soft correspondences. This allows correspondence learning across characters with length-variant keypoints, providing a stronger foundation for cross-category pose transfer.

\myparagraph{3D Pose Transfer.} 
Deformation transfer, {\ie}, the process of retargeting and transferring 3D poses across characters, is essential for animation, simulation, and virtual content creation.~\cite{yifan2020neural,yu2025mesh2animation,song2025puppeteer,wang2024towards,zhao2024pose,hong2025asmr,li2021learning,cha2025neural}.  
Early works~\cite{Gleicher_1998,sumner2004deformation,ben2009spatial,baran2009semantic} formulate deformation transfer as an optimization problem, but require handcrafted correspondences ({\eg}, point- or pose-wise) that are costly and non-scalable. Recent motion transfer methods~\cite{chen2025motion2motion,li2023ace} suffer from a reliance on exemplar motions from the target character, which constraints their applicability when such data is unavailable.  
Skeleton- or handle-based models~\cite{liao2022skeleton,chen2023weakly} predict relative transformations between keypoints, but typically assume one-to-one correspondences and are trained on pose datasets with limited human characters~\cite{mahmood2019amass,mixamo2025,li20214dcomplete}. As a result, they struggle to generalize across categories with inherently different skeletal structures. To bypass explicit correspondence supervision, recent works adopt implicit deformation fields~\cite{aigerman2022neural}, adversarial learning~\cite{chen2024towards,chen2021intrinsic}, cycle-consistency training~\cite{zhou2020unsupervised,gao2018automatic}, or conditional normalization layers~\cite{wang2020neural,chen2023lart}. Nevertheless, these methods remain restricted to structurally similar characters and fail on stylized or cross-species transfers.  

Recent novel large-scale articulation datasets~\cite{deitke2023objaverse,deng2025anymate,song2025magicarticulate} have enabled learning high-quality geometric features and reliable skeleton predictions~\cite{song2025magicarticulate} across diverse species. 
Drawing on these insights, we leverage pretrained shape encoders alongside a novel dataset of diverse animations to propose a cascade-transformer framework that extends pose transfer beyond humanoids to a broad range of characters.

%% file: 3-method.tex
\section{Methodology}

\subsection{Preliminary}

\myparagraph{Task Definition.}
Given a source character in canonical pose $\mathbf{\bar{V}}^{\text{src}} \in \mathbb{R}^{N^{\text{src}}\times 3}$, its posed instance $\mathbf{V}^{\text{src}}$ with pose parameters $\mathbf{p}$, and a target character in canonical pose $\mathbf{\bar{V}}^{\text{tgt}} \in \mathbb{R}^{N^{\text{tgt}}\times 3}$, the objective of pose transfer is to generate the posed target mesh $\mathbf{\hat{V}}^{\text{tgt}}$. Formally, we seek a mapping
$f(\mathbf{V}^{\text{src}}, \mathbf{\bar{V}}^{\text{src}}, \mathbf{\bar{V}}^{\text{tgt}}) \to 
\mathbf{\hat{V}}^{\text{tgt}}$,
where $\mathbf{\hat{V}}^{\text{tgt}}$ retains the target's geometry while transferring the pose $\mathbf{p}$ from the source.

\myparagraph{Character Articulation Formulation.}
We follow~\cite{liao2022skeleton,chen2023weakly} to represent character deformations via keypoints and skin weights.  
The keypoints $\mathbf{C}\in \mathbb{R}^{K\times 3}$ together with per-vertex skin weights $\mathbf{W} \in \mathbb{R}^{N \times K}$ are estimated by pretrained model~\cite{xu2020rignet}.
The number of keypoints $K$ may vary across characters (denoted as $\nkptsrc$ and $\nkpttgt$ for source and target characters).
The deformation of a character is achieved by assigning per-keypoint transformation $\mathbf{T} \in \mathbb{R}^{K\times7}$, and applying linear blend skinning (LBS)~\cite{kavan2014part} to deform the mesh from canonical space to posed space:
\vspace{-1ex}
\begin{equation}
\scalebox{1}{$
    \mathbf{v}_i = \sum\nolimits_{k=1}^{K} \mathbf{w}_{i,k} \, \mathbf{T}_{k} \bigl(\mathbf{\bar{v}}_i - \bar{\mathbf{c}}_k \bigr), \quad \forall \;\; \mathbf{\bar{v}}_i\in \bar{\mathbf{V}},$}
    \label{eq.lbs}
\end{equation}
where $\bar{\mathbf{c}}_k \in \bar{\mathbf{C}}$ denotes the canonical keypoints. For the posed character, the corresponding keypoints $\mathbf{c}_k$ are approximated by the weighted average of the deformed vertices.

\myparagraph{Geometry Priors.}  
We utilize a pretrained 3D shape encoder $\mathcal{E}$~\cite{zhao2023michelangelo} to extract geometry features $\mathbf{f}_{\mathcal{E}} \in \mathbb{R}^{l_{\mathcal{E}}\times d}$ from a given character mesh, where $l_{\mathcal{E}}$ denotes the number of query tokens.  
We extract geometry features for the canonical source ($\mathbf{f}_{\bar{\mathbf{V}}^{\text{src}}}$), posed source ($\mathbf{f}_{{\mathbf{V}}^{\text{src}}}$), and canonical target characters ($\mathbf{f}_{\bar{\mathbf{V}}^{\text{tgt}}}$). 
In addition, we define the residual geometry feature of the source character as $\delta_\mathbf{f} = \mathbf{f}_{{\mathbf{V}}^{\text{src}}} - \mathbf{f}_{\bar{\mathbf{V}}^{\text{src}}}$,
which captures the deformation difference between the posed and canonical source meshes in latent space.

\subsection{Motivation}  
\myparagraph{Challenges.} Humans possess the cognitive flexibility to imagine how animals imitate human behaviors and vice versa. In contrast, most existing pose transfer approaches remain restricted to narrow settings ({\eg}, humanoid-to-humanoid). A more general and practical goal is \emph{category-free pose transfer} across diverse characters. 
However, it is substantially more challenging, with two key obstacles:
\begin{enumerate}
\item \textbf{Diverse skeletons and geometries.}  
Characters differ drastically in skeletal structure, geometric topology, and mesh density. Directly estimating per-vertex correspondences across such variations is infeasible. A common workaround is to employ intermediate representations such as keypoints. 
Yet, existing works assume a fixed set of keypoints across categories, overlooks the inherent variation in anatomical structures ({\eg}, bones, limbs, and wings).
This mismatch makes one-to-one keypoint mapping unreliable and often degrades transfer quality.  

\item \textbf{Scarcity of paired data.}  
Datasets with diverse characters and artist-designed poses are extremely limited. As a result, most existing pose transfer methods rely on reference poses from large-scale humanoid datasets with only a few animal samples, which prevents them from generalizing to non-human characters.  
Moreover, the absence of ground-truth keypoint correspondences makes it difficult for models to learn accurate mappings in a purely self-supervised manner. Creating such correspondences manually is prohibitively costly and labor-intensive.
\end{enumerate}

\input{figure/fig_pipeline}

\myparagraph{Overview.}  
To tackle these challenges, we collect a new dataset of high-quality skeletons and diverse character motions spanning humanoids, quadrupeds, birds, and more. 
Building on this, we introduce {\name}, a cascade-transformer architecture designed for category-free pose transfer. {\name} learns soft correspondences across keypoints of different meshes and transfers poses accordingly.
An overview of our framework is illustrated in Fig.~\ref{fig.pipeline}, and the following sections detail our technical contributions.

\subsection{Data Collection} \label{sec.poseprior}

\myparagraph{Collecting Diverse Character Motions.}
To overcome the above challenges and enable generalization across character categories, we curate a new dataset, {\dataname}, from the web consisting of high-quality, artist-designed motions for a wide range of characters. 
These motions span various actions, such as running, sleeping, eating, and fighting.

For efficient training, we follow prior works~\cite{xu2020rignet,deng2025anymate,liao2022skeleton} to resample each character mesh into $5{,}000$ faces and recompute skinning weights via barycentric interpolation based on the artist-provided per-vertex weights. 
We additionally record bone names for each character, which provide structural cues in the text domain and enable connections across categories ({\eg}, the label ``limbs'' can correspond to ``arms'' for humanoids and ``wings'' for birds).  
All meshes are unified into \textit{.obj} format, while skeletal animations are stored in \textit{.bvh} format for consistency.

\myparagraph{Dataset Statistics.}
Tab.~\ref{tab.datacmp} summarizes {\dataname} in comparison with existing publicly available animation datasets.
Unlike prior collections, our dataset comprises $975$ characters spanning a broad spectrum of species and morphologies, including humanoids, quadrupeds, birds, reptiles, fishes, and insects.
The number of skeletal joints ranges from $11$ to $241$, with an average of $83$.
Each character is paired with artist-designed skeletal animations, resulting in a total of $28{,}809$ motions and $4{,}473{,}481$ frames, which is comparable in scale to the widely-used human motion datasets~\cite{mahmood2019amass}.
Examples that highlight the diversity and quality of our dataset are illustrated in Fig.~\ref{fig.dataset}.

\input{figure/fig_dataset}

\myparagraph{Pose Prior.}
Based on the large-scale pose dataset, we train a transformer-based pose prior model $\mathcal{F}$.
The model learns the distribution of plausible keypoint rotations sampled from the dataset across diverse characters, and model each rotation as a matrix-Fisher distribution~\cite{yin2022fishermatch,sengupta2021hierarchical,mohlin2020probabilistic}.
The trained $\mathcal{F}$ predicts per-keypoint distribution parameters $\hat{\mathbf{F}}$ for given poses, providing a statistical prior that (1) regularizes the rotation space during pose transfer training, and (2) helps ensure the generated transformations follow realistic, physically consistent patterns across different character morphologies.
Details of $\mathcal{F}$ can be found in Appendix.

\subsection{Cascade-Transformer for 3D Pose Transfer}

To achieve category-free pose transfer, we propose a cascade-transformer model, {\name}, that learns soft keypoint correspondences for shape-aware deformations and is trained with a two-stage scheme.

\myparagraph{Shape-aware Keypoint Correspondence.}
% \label{sec.corresptransformer}
We design a \emph{correspondence transformer} $\mathcal{G}$ that integrates shape conditioning to estimate soft correspondences between keypoint pairs of varying lengths.
Given canonical keypoints $\bar{\mathbf{C}}^{\text{src}}$ and $\bar{\mathbf{C}}^{\text{tgt}}$ from the source and target characters, respectively, $\mathcal{G}$ predicts a similarity matrix $\mathbf{S} \in \mathbb{R}^{\nkptsrc \times \nkpttgt}$ and its normalized counterpart, a doubly-stochastic matrix $\mathbf{M}$, representing the probabilistic correspondence between keypoints.
An overview of the $\mathcal{G}$ architecture is shown in Fig.~\ref{fig.gm}.

\input{figure/fig_gm}

In contrast to GNN-based models~\cite{wang2020combinatorial,wang2019learning} that rely on skeletal connectivity priors and may not generalize well to characters with diverse hierarchical structure, we directly encode spatial coordinates of keypoints using MLP layers to obtain keypoint tokens $g_{\mathbf{C}} \in \mathbb{R}^{K \times d_c}$.
To enhance discrimination between correlated body parts, we further incorporate shape features $\mathbf{f}_{\bar{\mathbf{V}}^\text{src}}$ and $\mathbf{f}_{\bar{\mathbf{V}}^\text{tgt}}$ extracted from pretrained models~\cite{zhao2023michelangelo}. These features are concatenated and processed by MLP layers to produce high-dimensional shape tokens $g_{\mathbf{M}} \in \mathbb{R}^{l_{\mathcal{E}} \times d_c}$.
The shape tokens are then combined with the keypoint tokens to form $[g_{\mathbf{M}}, g_{\mathbf{C}}]$, which are fed into transformer blocks to learn shape-aware latent representations $\mathbf{g}^{\text{src}}$ and $\mathbf{g}^{\text{tgt}}$ for the source and target keypoints.

The pairwise similarity matrix $\mathbf{S}$ between the source and target latent features is computed using a learnable affinity matrix $\mathbf{A}$.
The resulting similarity scores are exponentiated and normalized into a doubly stochastic correspondence matrix $\mathbf{M}$ via the Sinkhorn algorithm~\cite{sinkhorn1967concerning}, where each entry $\mathbf{M}_{i,j}$ represents the soft matching probability between a source and a target keypoint:
\begin{equation}
\resizebox{0.8\columnwidth}{!}{%
$\mathbf{S} = \exp \big( {\mathbf{g}^{\text{src}}}^{\top} \mathbf{A} \mathbf{g}^{\text{tgt}} \big), \quad
\mathbf{M} = \text{Sinkhorn}(\mathbf{S}).$}
\label{eq.affin}
\end{equation}

\myparagraph{Correspondence-aware Transformation Initialization.}
Given the canonical and posed meshes, keypoints, and skin weights of the source character, we first follow~\cite{besl1992method,liao2022skeleton} to estimate per-keypoint transformations {\small $\mathbf{T}^{\text{src}} = \{\mathbf{T}^{\text{src}}_1, \cdots, \mathbf{T}^{\text{src}}_{\nkptsrc}\}$}, where each transformation $\mathbf{T}^{\text{src}}_i$ includes a rotation quaternion $\mathbf{q}_i$ and a translation vector $\mathbf{t}_i$.

Using the correspondence matrix $\mathbf{M}$ from Eq.~\ref{eq.affin}, the initial transformations of the target character are obtained by aggregating the source transformations according to the matching probabilities.
For each target keypoint $\mathbf{c}_j \in \mathbf{C}^{\text{tgt}}$, its translation $\bar{\mathbf{t}}_j$ and associated query position $\bar{\mathbf{c}}_j$ in the source character are calculated as weighted averages:
\vspace{-1ex} 
\begin{equation}
    \resizebox{0.9\columnwidth}{!}{%
    $\bar{\mathbf{x}}_j = \left(\sum\nolimits_{i=1}^{\nkptsrc} \mathbf{M}_{i,j}\right)^{-1}
    \sum\nolimits_{i=1}^{\nkptsrc} \mathbf{x}_i \mathbf{M}_{i,j}, 
    \;\; \mathbf{x}_i \in \{\mathbf{t}_i, \mathbf{c}_i\}.$
    }
    \label{eq.transavg}
\end{equation}

\input{figure/fig_transfer}

Directly averaging quaternions, however, is invalid because it neither guarantees unit-norm rotations nor resolves the inherent $2\!:\!1$ ambiguity of quaternions--often leading to inconsistent or flipped orientations (see Fig.~\ref{fig.abl}).
Instead, we compute the weighted average rotation by minimizing the Frobenius norm between attitude matrices:
\vspace{-1ex}
\begin{equation}
    \resizebox{0.9\columnwidth}{!}{%
    $
\bar{\mathbf{q}}_j =
\argmin_{\mathbf{q}_j \in \mathbb{S}^3}
\sum\nolimits_{i=1}^{\nkptsrc}
\mathbf{M}_{i,j}
\| A(\mathbf{q}_j) - A(\mathbf{q}_i) \|_F^2,
$}
\label{eq.rotavg}
\end{equation}
where $A(\mathbf{q}) \in \mathrm{SO}(3)$ denotes the attitude matrix of quaternion $\mathbf{q}$.
Following~\cite{markley2007averaging,oshman2006attitude}, the solution $\bar{\mathbf{q}}_j$ is given by the eigenvector corresponding to the largest eigenvalue of the weighted covariance matrix, {\ie}, {\small$\sum\nolimits_{i=1}^{\nkptsrc} \mathbf{M}_{i,j}\, \mathbf{q}_i \mathbf{q}_i^{\top}$}.

\myparagraph{Shape-aware Pose Transfer.}
Given the initialized transformations, our goal is to refine them into the final target transformations while incorporating geometric conditions.
To achieve this, we design a \emph{pose transfer transformer} $\mathcal{H}$ that maps shape and keypoint features to per-keypoint transformations of the target character, as illustrated in Fig.~\ref{fig.transfer}.

We extract geometry features $\mathbf{f}_{\bar{\mathbf{V}}^{\text{src}}}, 
\mathbf{f}_{\mathbf{V}^{\text{src}}}, \mathbf{f}_{\bar{\mathbf{V}}^{\text{tgt}}}$ 
from the canonical and posed source meshes and the canonical target mesh, respectively. 
These features are fused through cross-attention layers to inject the residual source deformation cues $\delta_\mathbf{f}$ into the target representation, producing geometry-condition tokens $h_{\mathbf{M}}$.
For the keypoint representation, each target keypoint $\mathbf{c}_j$ is paired with its query position $\bar{\mathbf{c}}_j$ and initialized transformation ${\bar{\mathbf{t}}_j, \bar{\mathbf{q}}_j}$.
The concatenated vector $[\mathbf{c}_j, \bar{\mathbf{c}}_j, \bar{\mathbf{t}}_j, \bar{\mathbf{q}}_j]$ is then projected through MLP layers to form high-dimensional keypoint tokens $h_{\mathbf{C}} \in \mathbb{R}^{\nkpttgt \times d_c}$.

Finally, the geometry tokens $h_{\mathbf{M}}$ and keypoint tokens $h_{\mathbf{C}}$ are concatenated to form $[h_{\mathbf{M}}, h_{\mathbf{C}}]$ and fed into transformer blocks, producing shape-aware latent features.
MLP layers decode these features into per-keypoint transformations {\small $\mathbf{T}^{\text{tgt}} = \{\hat{\mathbf{T}}^{\text{tgt}}_1, \ldots, \hat{\mathbf{T}}^{\text{tgt}}_{\nkpttgt}\}$}, which are applied to deform the canonical target mesh into final posed mesh $\hat{\mathbf{V}}^{\text{tgt}}$ via Eq.~\ref{eq.lbs}.

\subsection{Training, Inference \& Objective Functions}

\myparagraph{Text-Guided Ground-truth Correspondence.}
A key prerequisite for correspondence learning is access to reliable ``ground-truth'' matches.  
However, manually annotating either keypoint- or vertex-level correspondences for large-scale pairs is highly non-trivial and labor-intensive.  

Fortunately, the artist-designed characters themselves furnish semantic keypoint names. For example, the arms of humanoids and the wings of birds are both labeled as ``limbs'', which aligns with human perception of their functional similarity. Drawing from this observation, we bypass handcrafting correspondences or designing sophisticated algorithms~\cite{zhan2024charactermixer,xu2022hierarchical}.
Instead, we use CLIP $\mathcal{E}_{\text{CLIP}}$~\cite{radford2021learning} to encode the textual labels of keypoints $\mathbf{c}_k$ into latent space, yielding $\mathbf{f}_k$. 
We compute the similarity matrix $\mathbf{S}_{\cos} \in \mathbb{R}^{{\nkptsrc}\times {\nkpttgt}}$ via cosine similarity, each element $\mathbf{s}_{i,j}$ given by:  
\vspace{-1ex}
\begin{equation}
\mathbf{s}_{i,j} = \frac{\mathbf{f}_i \cdot \mathbf{f}_j}{\|\mathbf{f}_i\| \, \|\mathbf{f}_j\|},
\label{eq.textsim}
\end{equation}
where $\mathbf{s}_{i,j}$ measures the similarity between source and target characters $\mathbf{c}_i \in \mathbf{C}^{\text{src}}$ and $\mathbf{c}_j \in \mathbf{C}^{\text{tgt}}$.
Finally, we normalize $\mathbf{S}_{\cos}$ using the Hungarian algorithm~\cite{kuhn1955hungarian} to obtain ground-truth one-hot correspondence $\mathbf{M}_{\text{hung}}$. Fig.~\ref{fig.corresp} showcase examples of our text-based ground-truth {\vs} hierarchical correspondence~\cite{zhan2024charactermixer}.  

\input{figure/fig_corresp}

\myparagraph{Two-Stage Training.}
We train {\name} using two-stage process. 
In the first stage, we train the correspondence transformer $\mathcal{G}$. We note that multiple keypoints from the source character may correspond to a single keypoint in the target, and vice versa.
Therefore, in addition to the hard one-to-one mapping $\mathbf{M}_{\text{hung}}$ obtained via the Hungarian algorithm, we introduce a soft-matching matrix $\mathbf{M}_{\text{sink}} = \text{Sinkhorn}(\mathbf{S}_{\cos})$, which captures many-to-many correlations between source and target keypoints.
We then use the Frobenius norm to jointly encourage the predicted affinity matrix $\mathbf{S}$ to align with the text-based cosine similarity matrix $\mathbf{S}_{\cos}$, while maintaining consistency between the predicted correspondence $\mathbf{M}$, the soft Sinkhorn mapping $\mathbf{M}_{\text{sink}}$, and the hard assignment $\mathbf{M}_{\text{hung}}$:
\vspace{-1ex}
\begin{equation}
\resizebox{0.95\columnwidth}{!}{$
    \mathcal{L}_{\text{forb}} = \big\|\mathbf{S} - \mathbf{S}_{\cos} \big\|_2^2 + \big\|\mathbf{M} - \mathbf{M}_{\text{sink}}\big\|_2^2 + \big\|\mathbf{M} - \mathbf{M}_{\text{hung}}\big\|_2^2.$}
\end{equation}

In the second stage, we freeze the correspondence transformer $\mathcal{G}$ and train the pose transfer transformer $\mathcal{H}$ using a cycle-consistency objective.
Since ground-truth targets are unavailable for cross-category pairs, we adopt a self-supervised strategy that reconstructs the source character from its transferred pose.
Specifically, we first transfer the pose from the source to the target, and then back from the predicted target to the source. The reconstruction loss encourages the recovered source mesh to match the given one:
\vspace{-1.5ex}
\begin{equation}
\resizebox{0.4\columnwidth}{!}{$
    \mathcal{L}_{\text{rec}} = \big\| \hat{\mathbf{V}}^{\text{src}} - \mathbf{V}^{\text{src}} \big\|_2^2, $}
\label{eq.cycleloss}
\end{equation}
where 
{\small $\mathbf{\hat{V}}^{\text{tgt}} = f(\mathbf{V}^{\text{src}},       \mathbf{\bar{V}}^{\text{src}}, \mathbf{\bar{V}}^{\text{tgt}})$} and 
{\small $\mathbf{\hat{V}}^{\text{src}} = f(\mathbf{\hat{V}}^{\text{tgt}}, \mathbf{\bar{V}}^{\text{tgt}}, \mathbf{\bar{V}}^{\text{src}})$}.

To further regularize rotations, we employ the pretrained pose prior model $\mathcal{F}$.
Given the ground-truth posed character $\mathbf{V}^{\text{src}}$, we use $\mathcal{F}$ to estimate the pose distribution parameters $\hat{\mathbf{F}}$ of each keypoints. Then, if the predicted rotation $\hat{\mathbf{q}}_{k}$ for $k$-th keypoint is reasonable, we assume it shows the max $\log$-likelihood given the distribution parameters based on $\mathbf{V}^{\text{src}}$. Hence, we enforce the estimated rotation to preserve minimal negative log-likelihood (NLL) properties~\cite{sengupta2021hierarchical,mohlin2020probabilistic}.
\vspace{-1ex}
\begin{equation}
\resizebox{0.7\columnwidth}{!}{$
\begin{aligned}
    \mathcal{L}_{\text{reg}} 
    & = - \sum\nolimits_{k=1}^{K} \log p\left( A(\hat{\mathbf{q}}_{k}) \mid \hat{\mathbf{F}}_k \right) \\
    & = \sum\nolimits_{k=1}^{K} 
    \left( 
    \log c(\hat{\mathbf{F}}_k) 
    - \operatorname{tr}\left( 
    \hat{\mathbf{F}}_k^\top A(\hat{\mathbf{q}}_{k}) 
    \right) 
    \right)
\end{aligned},
$}
\label{eq.priorloss}
\end{equation}
where $\hat{\mathbf{F}}_k$ denotes the predicted distribution parameters for the $k$-th keypoint, $c(\hat{\mathbf{F}}_k)$ is the normalization constant {\wrt} the distribution parameters $\hat{\mathbf{F}}_k$.

Additionally, we encourage the reconstructed meshes in the cycle process to preserve consistent high-level geometric features with their original posed counterparts by enforcing feature-space consistency, {\small $\mathcal{L}_{\text{feat}} =\big\| \mathcal{E}(\hat{\mathbf{V}}^{\text{src}}) - \mathcal{E}({\mathbf{V}}^{\text{src}}) \big\|_2^2$}.

\input{figure/fig_main_cmp}

\myparagraph{Inference Stage.}
During inference, given the canonical poses of both the source and target characters, we first apply the pretrained skeleton prediction model~\cite{xu2020rignet} to obtain their corresponding skin weights and skeleton structures.
{\name} then deforms the source pose into the target character according to the learned soft correspondences and transformation mappings.
Finally, following prior works~\cite{liao2022skeleton,zhang2024tapmo}, we perform test-time refinement using an as-rigid-as-possible (ARAP)~\cite{sorkine2007rigid} optimization to enhance mesh smoothness and preserve local geometric details in the final deformed results.

%% file: figure/fig_pipeline.tex
\begin{figure}
    \centering
    \begin{overpic}[trim=1.2cm 0cm 0cm 0cm,clip,width=\linewidth,grid=false]{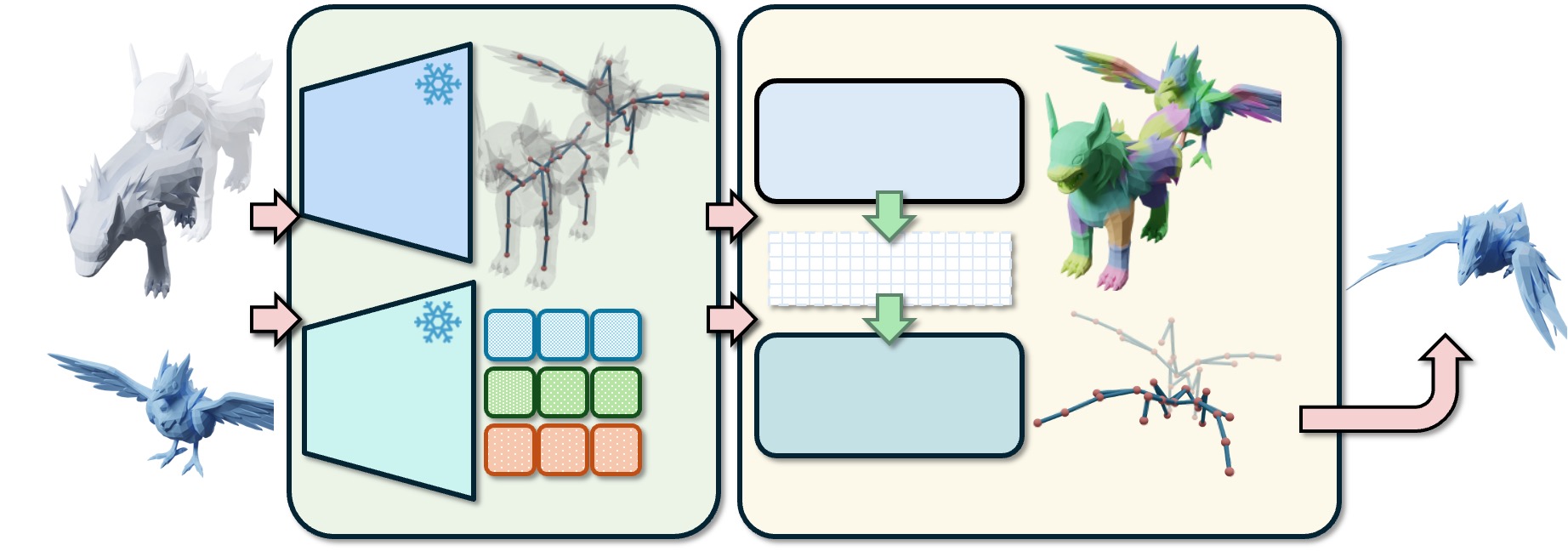}
    \put(0,3){\tiny \textbf{Target Character}}
    \put(2.5,15){\tiny \textbf{Source Pose}}
    \put(21.5,26.5){\tiny {Skin}}
    \put(19.5,24.5){\tiny {Predictor}}
    \put(21,11){\tiny {Shape}}
    \put(20,9){\tiny {Encoder}}
    \put(34,18.5){\tiny {Keypoints \&}}
    \put(33,16.5){\tiny {Skin Weights}}
    \put(30,2){\tiny {Shape Features}}
    \put(40.2,13.5){{$\Scale[0.6]{\mathbf{f}_{\bar{\mathbf{V}}^{\text{src}}}}$}}
    \put(40.2,9.5){{$\Scale[0.6]{\mathbf{f}_{{\mathbf{V}}^{\text{src}}}}$}}
    \put(40.2,5.5){{$\Scale[0.6]{\mathbf{f}_{\bar{\mathbf{V}}^{\text{tgt}}}}$}}

    \put(49,27){\tiny {Correspondence}}
    \put(50,25){\tiny {Transformer $\mathcal{G}$}}
    \put(49.5,10.5){\tiny {Pose Transfer}}
    \put(49,8.5){\tiny {Transformer $\mathcal{H}$}}
    \put(49.5,18.75){\tiny \textit{Transformation}}
    \put(51,16.75){\tiny \textit{Initialization}}
    \put(50,33){\tiny {\ul{Stage I}: Correspondence Pretraining}}
    \put(50,2){\tiny {\ul{Stage II}: Cycle-consistency Training}}
    \put(64,4.5){\tiny \textbf{Target Transformation}}
    \put(71.5,18.5){\tiny \textbf{Soft-Matching}}
    \put(69,16){\tiny \textbf{ Correspondence}}
    \put(87,14.5){\tiny \textbf{Target Pose}}
    \put(84,11.5){\tiny + Eq.~\ref{eq.lbs}}
    \end{overpic}
    \vspace{-1.5em}
    \caption{\textbf{Overview of {\name} for category-free pose transfer.}
    {\name} takes a paired source pose and target character as input.
    It first employs the correspondence transformer $\mathcal{G}$ to estimate soft keypoint correspondences, then refines the initialized transformations using the pose transfer transformer $\mathcal{H}$ to generate the target transformations.
    Finally, the target character is deformed into the desired pose through linear blend skinning (LBS).
    }
    \label{fig.pipeline}
        \vspace{-15pt}
\end{figure}

%% file: figure/fig_dataset.tex
\begin{figure}[t!]
    \centering
    \begin{overpic}[trim=1cm 5cm 1cm 5.5cm,clip,width=\linewidth,grid=false]{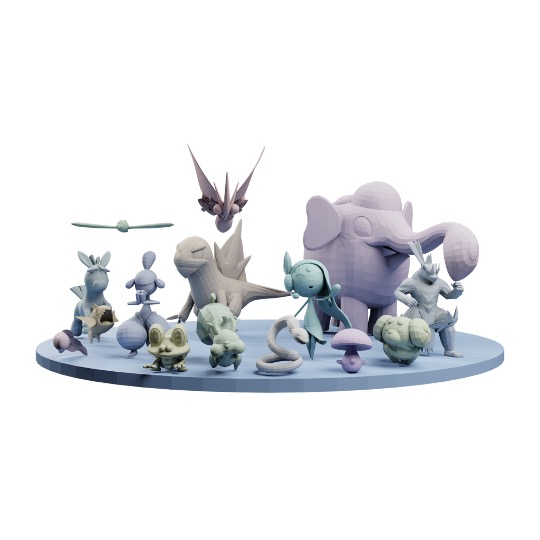}\end{overpic}
    \vspace{-2em}
    \caption{\textbf{Pose examples from the {\dataname}.} Our dataset covers a wide range of species (including humanoids, insects, quadrupeds, fishes, {\etc}) with high-quality, artist-designed poses. 
    }
    \label{fig.dataset}
    \vspace{-15pt}
\end{figure}

%% file: figure/fig_gm.tex
\begin{figure}
    \centering
    \begin{overpic}[trim=0.cm 0cm 0cm 0cm,clip,width=\linewidth,grid=false]{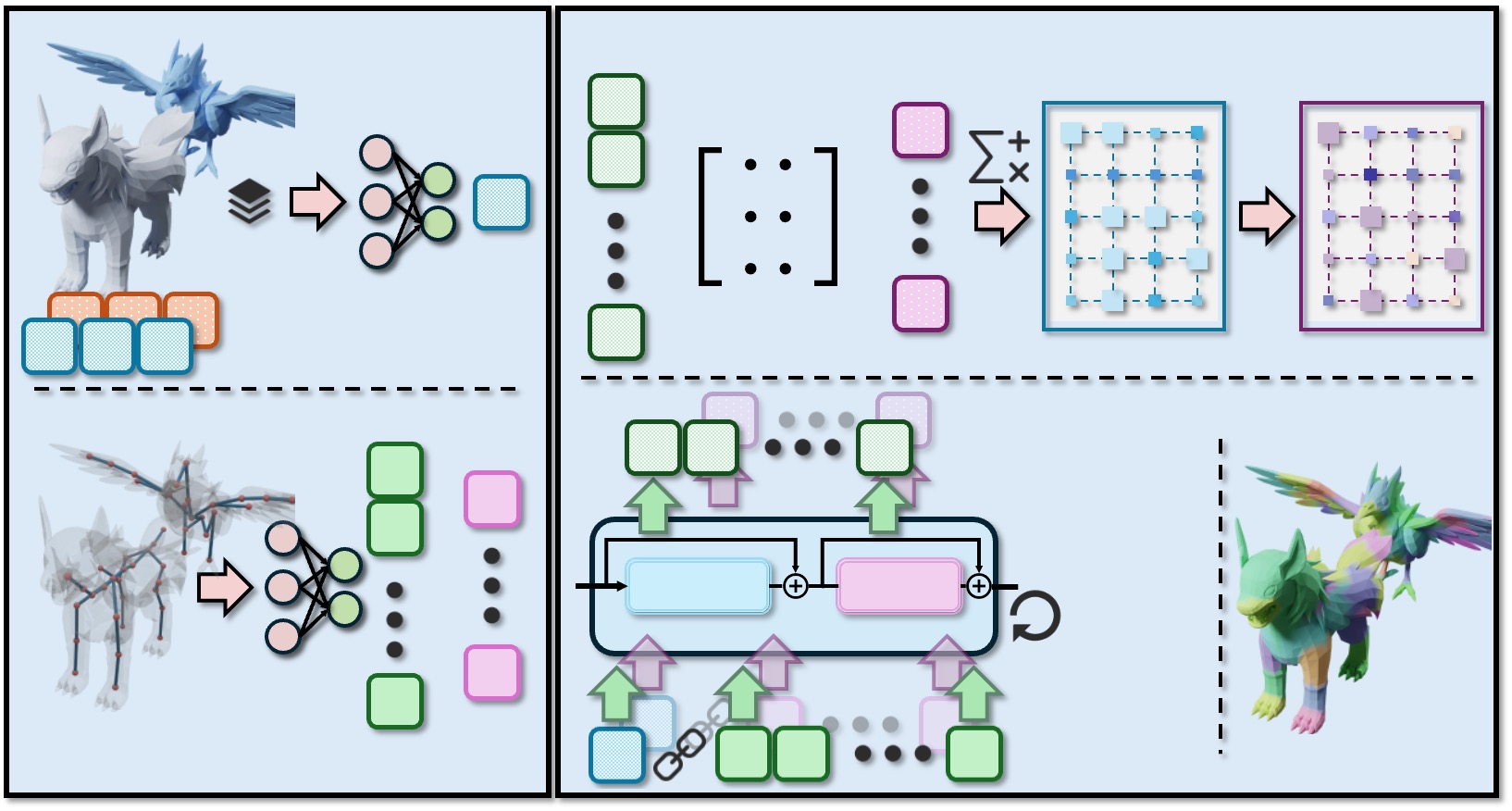}
    \put(3,6){\tiny Source/Target}
    \put(4,4){\tiny Keypoints}
    
    \put(15.8,8.5){\tiny \textit{Keypoint}}
    \put(16,6.5){\tiny \textit{Encoder}}

    \put(21.5,25){\tiny Keypoint Tokens}
    \put(23.5,2.8){ $\Scale[0.6]{{g_{\mathbf{C}}^{\text{src}}}}$}
    \put(31,4.5){$\Scale[0.6]{{g_{\mathbf{C}}^{\text{tgt}}}}$}
    \put(16,32.5){$\Scale[0.6]{\mathbf{f}_{\bar{\mathbf{V}}^\text{tgt}}}$}
    \put(15,29.5){$\Scale[0.6]{\mathbf{f}_{\bar{\mathbf{V}}^\text{src}}}$}

    \put(25,34){\tiny \textit{Shape}}
    \put(23.5,32){\tiny \textit{Projector}}

    \put(30.5,45){\tiny Shape}
    \put(30,43){\tiny Tokens}
    \put(31.5,36){$\Scale[0.6]{{g_{\mathbf{M}}}}$}

    \put(3,51){\tiny \textbf{\ul{(a)}. Shape \& Keypoint Tokenization}}
    \put(44,14.3){\tiny Attn.}
    \put(57.5,14.3){\tiny MLP}
    \put(47,11){\tiny \textit{Transformer Blocks}}
    \put(69,8){\tiny Shape \& }
    \put(66,6){\tiny Keypoint Tokens}
    \put(67,4){\tiny (Source/Target)}
    \put(70,11){\tiny $\times$\textit{N}}
    \put(37.5,23.5){$\Scale[0.6]{\mathbf{g}^\text{src}}$}
    \put(43,26.5){$\Scale[0.6]{\mathbf{g}^\text{tgt}}$}

    \put(62,23){\tiny \textbf{\ul{(b)}. Shape-Keypoint}}
    \put(67,20.5){\tiny \textbf{Interaction}}

    \put(39,50){$\Scale[0.6]{\mathbf{g}^\text{src}}$}
    \put(59,48){$\Scale[0.6]{\mathbf{g}^\text{tgt}}$}

    \put(42.5,48){\tiny $\boldsymbol{{\top}}$}

    \put(46,48){\tiny \textit{Learnable}}
    \put(46,46){\tiny \textit{Weight} $\mathbf{A}$}

    \put(45,31){\tiny Affinity Matrix}

    \put(71,50){\tiny Similarity}
    \put(71,48){\tiny Matrix $\mathbf{S}$}

    \put(83,50){\tiny Doubly-Stochastic }
    \put(86,48){\tiny Matrix $\mathbf{M}$}

    \put(64,36){\tiny Eq.~\ref{eq.affin}}

    \put(82,36){\tiny \mbox{\cite{sinkhorn1967concerning}}}

    \put(70,30){\tiny \textbf{\ul{(c)}. Correspondence Estimation}}

    \put(85,3){\tiny \textbf{\ul{(d)}. Output}}
    \put(83.5,25){\tiny Soft-Matching}
    \put(83,23){\tiny Correspondence}

    \end{overpic}
    \vspace{-2em}
    \caption{
    \textbf{Overview of the correspondence transformer $\mathcal{G}$.}
    We (a) first extract shape and keypoint tokens using the shape projector and keypoint encoder,
    (b) fuse shape conditions with respective keypoint latents through transformer blocks,
    (c) estimate correspondences via learnable affinity weights followed by the Sinkhorn algorithm,
    and (d) produce soft-matching correspondences between the given characters.
    }
    \label{fig.gm}
        \vspace{-12pt}
\end{figure}
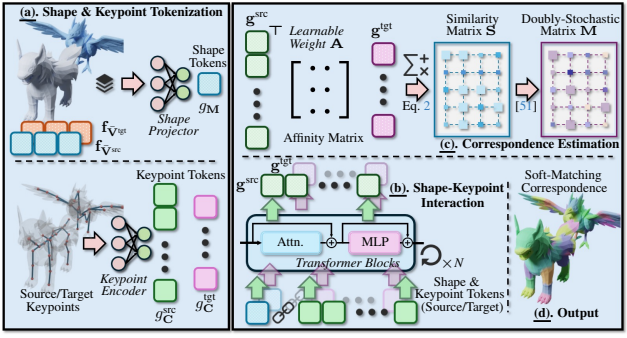

%% file: figure/fig_transfer.tex
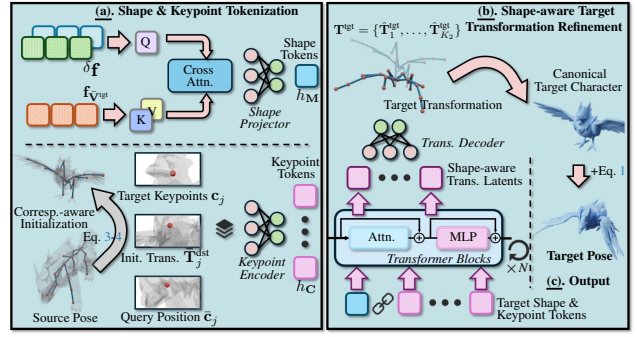
\begin{figure}[t!]
    \centering
    \begin{overpic}[trim=0.15cm 0cm 0cm 0cm,clip,width=\linewidth,grid=false]{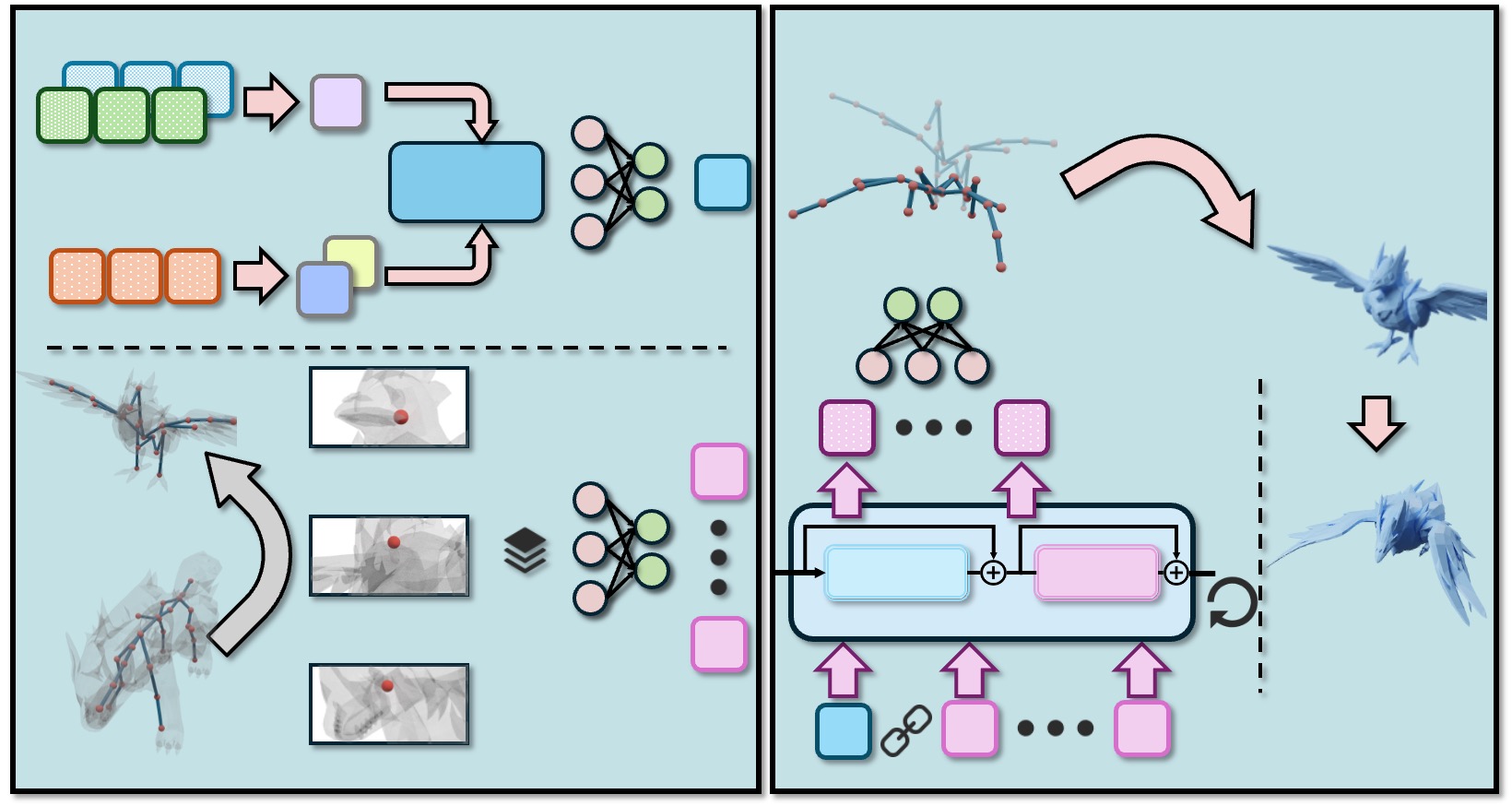}

    \put(14,51){\tiny \textbf{\ul{(a)}. Shape \& Keypoint Tokenization}}
    \put(12,42){$\Scale[0.6]\delta_\mathbf{f}$}
    \put(12,38){$\Scale[0.6]{\mathbf{f}_{\bar{\mathbf{V}}^{\text{tgt}}}}$}
    \put(21.2,46.2){\tiny Q}
    \put(20.5,33.6){\tiny K}
    \put(22.3,35){\tiny V}
    \put(28,41.7){\tiny Cross}
    \put(28.5,39.7){\tiny Attn.}

    \put(38.5,35){\tiny \textit{Shape} }
    \put(37,33){\tiny  \textit{Projector}}
    
    \put(44,46){\tiny Shape}
    \put(43.5,44){\tiny Tokens}
    
    \put(46,37.5){$\Scale[0.6]{h_{\mathbf{M}}}$}

    \put(1,19){\tiny Corresp.-aware}
    \put(2,17){\tiny Initialization}
    \put(4,2){\tiny Source Pose}

    \put(12,15){\tiny Eq.~\ref{eq.transavg}-\ref{eq.rotavg}}
    \put(18,2.2){{\tiny{Query Position }}$\Scale[0.6]{\bar{\mathbf{c}}_j}$}
    \put(18,11.7){{\tiny{Init. Trans. }}$\Scale[0.6]{\mathbf{{\bar{T}}}^{\text{dst}}_{j}}$}
    \put(17.5,21.5){{\tiny{Target Keypoints }}$\Scale[0.6]{\mathbf{c}_j}$}

    \put(36.8,11){\tiny \textit{Keypoint} }
    \put(37,9){\tiny  \textit{Encoder}}

    \put(42,27){\tiny Keypoint }
    \put(43,25){\tiny Tokens }

    \put(46,7){$\Scale[0.6]{h_{\mathbf{C}}}$}

    \put(52,48){$\Scale[0.5]{\mathbf{T}^{\text{tgt}} = \{\hat{\mathbf{T}}^{\text{tgt}}_1, \ldots, \hat{\mathbf{T}}^{\text{tgt}}_{\nkpttgt}\}}$}

    \put(75,51){\tiny \textbf{\ul{(b)}. Shape-aware Target }}
    \put(73,48.5){\tiny \textbf{Transformation Refinement}}

    \put(60,36){\tiny Target Transformation}
    \put(66,30){\tiny \textit{Trans. Decoder}}

    \put(86,12){\tiny \textbf{Target Pose}}
    \put(86,8){\tiny \textbf{\ul{(c)}. Output}}

    \put(93,25.5){\tiny +Eq.~\ref{eq.lbs}}

    \put(70.2,26){\tiny Shape-aware}
    \put(70,24){\tiny Trans. Latents}

    \put(87,41){\tiny Canonical}
    \put(84,39){\tiny Target Character}

    \put(79.2,10){\tiny $\Scale[1]{\times}$\textit{N}}

    \put(57.5 ,15){\tiny Attn.}
    \put(70.5,15){\tiny MLP}
    \put(60.5,11.5){\tiny \textit{Transformer Blocks}}

    \put(78,4.5){\tiny Target Shape \& }
    \put(78,2.5){\tiny Keypoint Tokens}
    \end{overpic}
    \vspace{-2em}
    \caption{
    \textbf{Overview of the pose transfer transformer $\mathcal{H}$.}
    We (a) first perform cross-attention to extract deformation-aware cues for shape tokenization and apply correspondence-aware initialization for keypoint tokenization.
    (b) The shape and keypoint tokens are fed into transformer blocks to derive high-level representations, and decode into refined target transformations.
    (c) the posed target mesh is generated by deforming the canonical target through Eq.~\ref{eq.lbs}.
    }
    \label{fig.transfer}
        \vspace{-12pt}
\end{figure}

%% file: figure/fig_corresp.tex
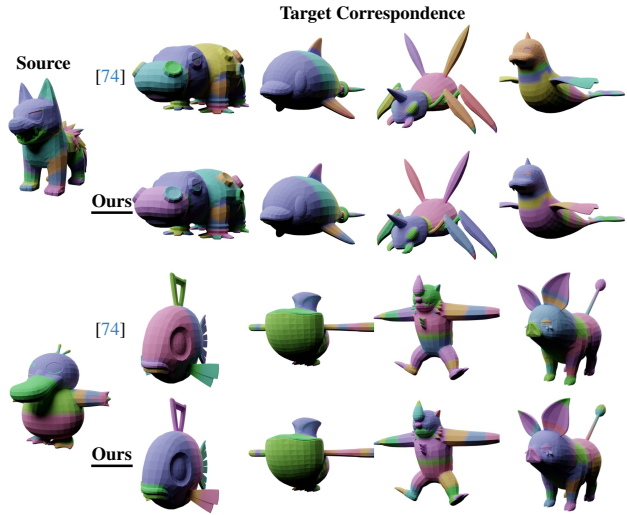
\begin{figure}[t!]
    \centering
    \begin{overpic}[trim=0cm 0cm 0cm 0cm,clip,width=\linewidth,grid=false]{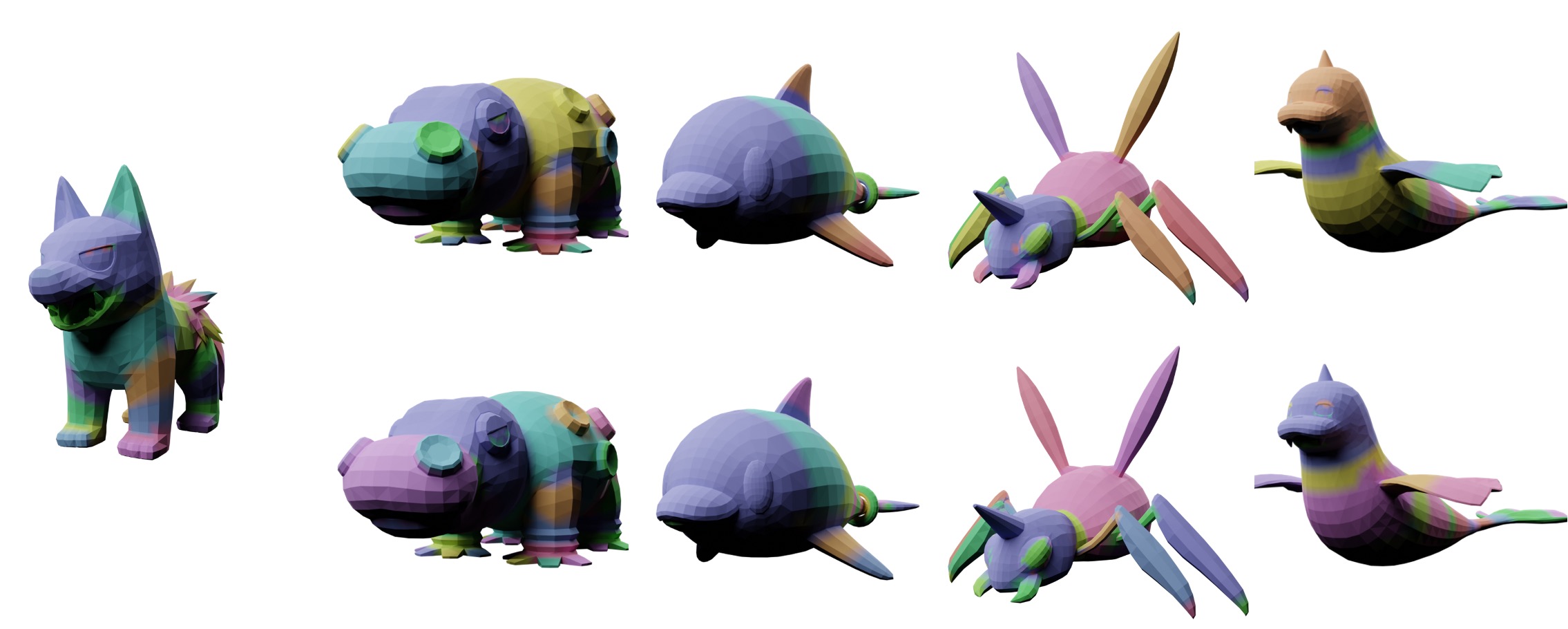}
    \put(15.5,29){\scriptsize \mbox{\cite{zhan2024charactermixer}}}
    \put(15,9){\scriptsize \textbf{\ul{Ours}}}
    \put(3,31){{\scriptsize \textbf{Source}}}
    \put(45,39){{\scriptsize \textbf{Target Correspondence}}}
    \end{overpic}
    \begin{overpic}[trim=0cm 0cm 0cm 0cm,clip,width=\linewidth,grid=false]{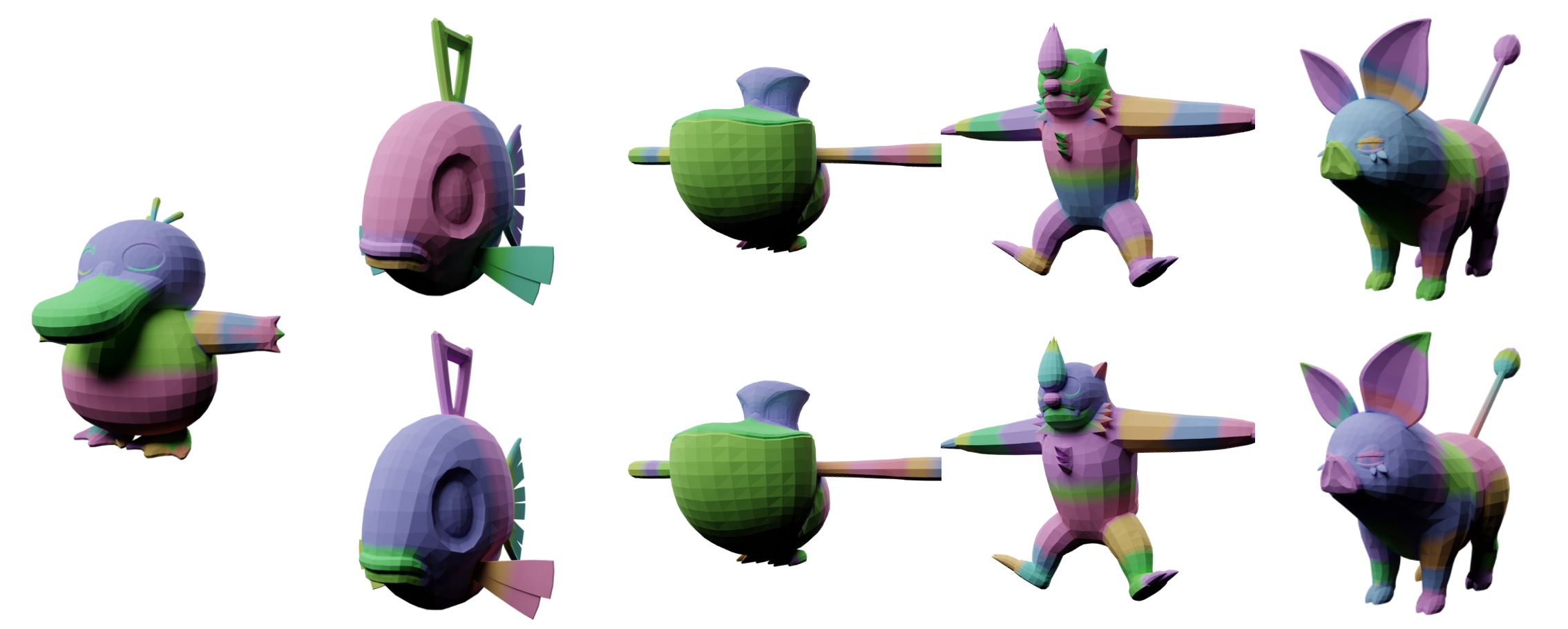}
    \put(15.5,29){\scriptsize \mbox{\cite{zhan2024charactermixer}}}
    \put(15,9){\scriptsize \textbf{\ul{Ours}}}
    \end{overpic}
    \vspace{-2em}
    \caption{\textbf{Correspondence visualization.}
    We visualize correspondences from source characters (\textit{left}) to category-free targets (\textit{right}). 
    Compared with the hierarchical correspondence algorithm~\cite{zhan2024charactermixer,xu2022hierarchical}, our text-guided correspondence yields more coherent and semantically consistent part alignments across characters.}
    \label{fig.corresp}
        \vspace{-10pt}
\end{figure}

%% file: figure/fig_main_cmp.tex
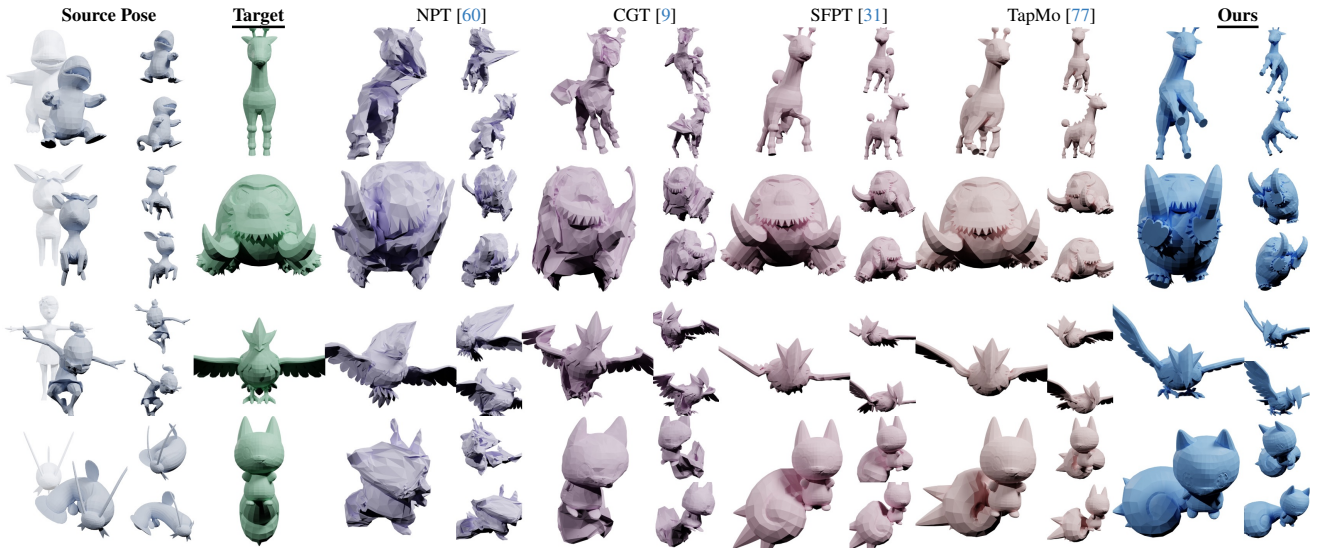
\begin{figure*}[ht!]
    \centering
    \begin{overpic}[trim=0cm 0cm 0cm -1cm,clip,width=\linewidth,grid=false]{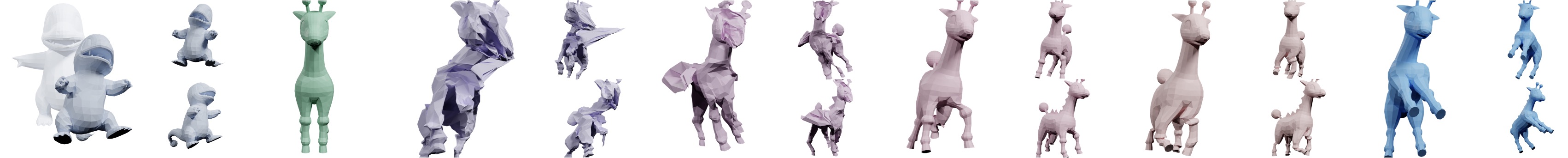}
    \put(5,10.5){\scriptsize \textbf{Source Pose}}
    \put(18,10.5){\scriptsize \textbf{\ul{Target}}}
    \put(32,10.5){\scriptsize \scriptsize \mbox{NPT~\cite{wang2020neural}}}
    \put(47,10.5){\scriptsize \scriptsize \mbox{CGT~\cite{chen2022geometry}}}
    \put(62,10.5){\scriptsize \scriptsize \mbox{SFPT~\cite{liao2022skeleton}}}
    \put(77,10.5){\scriptsize \scriptsize \mbox{TapMo~\cite{zhang2024tapmo}}}
    \put(93,10.5){\scriptsize \scriptsize \textbf{\ul{Ours}}}
    \end{overpic}
    \begin{overpic}[trim=0cm 0cm 0cm 0cm,clip,width=\linewidth,grid=false]{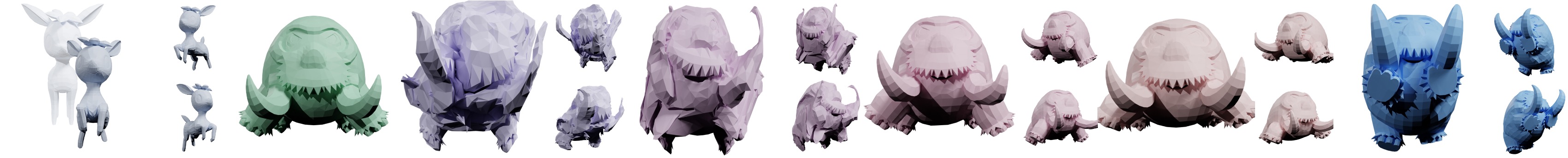}    
    \end{overpic}
    \begin{overpic}[trim=0cm 0cm 0cm 0cm,clip,width=\linewidth,grid=false]{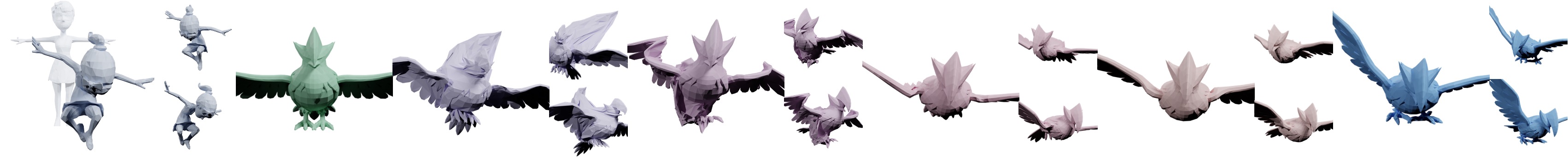}    
    \end{overpic}
    \begin{overpic}[trim=0cm 0cm 0cm 0cm,clip,width=\linewidth,grid=false]{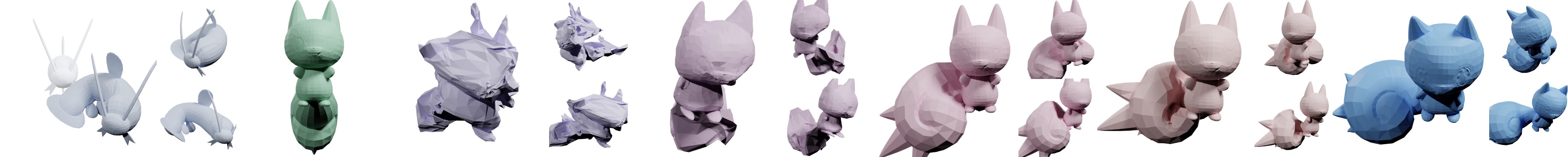}    
    \end{overpic}
    \vspace{-2em}
    \caption{
    \textbf{Qualitative comparisons with existing methods.}  
    We show pose transfer results across a wide range of character categories.  
    From \textit{left} to \textit{right}: source pose, target character, and results from different methods.  
    Each example is rendered from three viewpoints for comprehensive visual comparisons.  
    Additional qualitative results are provided in the Appendix.
    }
    \label{fig.mainres}
    \vspace{-15pt}
\end{figure*}

%% file: 4-exp.tex
\section{Experiments}
\label{sec.exp}

\subsection{Implementation Details}
\label{sec.impdetail}

We implement {\name} using PyTorch framework~\cite{paszke2019pytorch} and train it with the AdamW optimizer~\cite{loshchilov2018decoupled}, starting with an initial learning rate of $1e{-4}$ for both stages.
Our training corpus combines Mixamo~\cite{mixamo2025}, AMASS~\cite{mahmood2019amass}, and our newly collected {\dataname}, covering diverse and complex shapes and poses. The dataset is split into training, validation, and test sets.
In the first stage, we train the correspondence transformer with $384$k source-target pairs using a mini-batch size of $128$ for $60$ epochs.
In the second stage, we train the pose transfer transformer for $200$ epochs with a mini-batch size of $100$, sampling $100$k random poses per epoch.
All experiments are conducted on $8$ NVIDIA A100 GPUs.
More details are provided in the Appendix.

\subsection{Benchmark, Evaluation Metrics \& Baselines}

\myparagraph{Benchmarks.}
We evaluate category-free pose transfer under two settings.  
(1) \textit{Intra-category transfer:} We select the widely-use Mixamo~\cite{mixamo2025} to assess pose transfer among humanoid characters. Since all Mixamo characters share the same skeleton structure, the ground-truth targets are generated by directly applying the source transformations. We construct $100$ character pairs, each with $20$ random test poses, resulting in a total of $2{,}000$ evaluation cases.
(2) \textit{Cross-category transfer:} To assess generalization beyond specific categories, where ground-truth poses are unavailable, we adopt a cycle-consistency evaluation, where each method performs pose transfer in both directions, and the consistency between the source and reposed character is measured to quantify transfer quality. We randomly sample $300$ character pairs from our dataset and Mixamo, covering both humanoid-to-any and any-to-humanoid settings, with $10$ poses per pair, resulting in $3{,}000$ evaluation cases.

\myparagraph{Evaluation Metrics \& Baselines.}
Following previous works~\cite{wang2023zero,liao2022skeleton,yoo2024neural}, we adopt two metrics for quantitative evaluation:  
{Point-wise Mesh Euclidean Distance (PMD)}, which measures pose similarity between the predicted and ground-truth deformations, and  
{Edge Length Score (ELS)}, which evaluates edge consistency after deformation to assess the overall mesh smoothness. 
We compare {\name} against $4$ state-of-the-art pose transfer methods~\cite{liao2022skeleton,wang2020neural,chen2022geometry,zhang2024tapmo}.  
As {\name} is among the first to address category-free pose transfer, previous work was originally trained within specific domains. For fair comparison, we re-train or fine-tune their publicly available implementations using the same mixed-character training data described in Sec.~\ref{sec.impdetail}.

\subsection{Qualitative Comparisons}

In Fig.~\ref{fig.mainres}, we present qualitative results for category-free pose transfer across diverse character types, including humanoids, fishes, quadrupeds, and birds, demonstrating the robustness of {\name} compared to prior methods.  
Previous approaches ({\eg}, NPT~\cite{wang2020neural} and CGT~\cite{chen2022geometry}) rely on self-supervised learning within paired data of similar characters, limiting their ability to generalize to unseen categories.  
SFPT~\cite{liao2022skeleton} and TapMo~\cite{zhang2024tapmo} depend on a fixed number of handle points for deformation, which restricts them to one-to-one mappings and leads to artifacts or twisted poses when transferring across topologically different characters.  
In contrast, the proposed {\name} leverages soft keypoint correspondences for flexible many-to-many mappings, enabling natural and semantically consistent pose transfer across morphologically diverse characters.  
For instance, in the \textit{third row}, a human pose with arms fully extended is vividly transferred to a bird spreading its wings, while maintaining structural consistency such as the alignment between the human thighs and bird claws.

\subsection{Quantitative Comparisons}

Quantitative results are shown in Tab.~\ref{tab.cmp}, with the best results \textbf{bolded}.
NPT~\cite{wang2020neural} and CGT~\cite{chen2022geometry} are designed for transferring poses with same topology, therefore their performance degrades noticeably in cross-category scenarios.
SFPT~\cite{liao2022skeleton} and TapMo~\cite{zhang2024tapmo} show improved generalization in category-free scenarios but remain constrained by their one-to-one correspondence assumption.
In contrast, {\name} consistently achieves the best results across both settings, which demonstrate that our model effectively captures and transfers pose characteristics even across characters with distinct structures and topologies.
These gains stem from our shape-aware transformer design, which leverages geometric priors and flexible, length-variant keypoint correspondences to enable robust and accurate pose transfer.

\input{table/tab_cmp}

\subsection{Application of {\name}}

The category-free nature of {\name} enables it to serve as a plug-and-play module for text-to-any-character motion generation, greatly expanding the applicability of existing T2M systems.
As shown in Fig.~\ref{fig.t2m}, we take motions produced by off-the-shelf T2M models~\cite{jiang2023motiongpt,chen2023executing}--originally defined on SMPL skeletons~\cite{loper2015smpl,pavlakos2019expressive}--and use {\name} to transfer them frame-by-frame into a wide variety of target characters, yielding diverse and visually compelling animations.
This demonstrates the potential of {\name} as a general-purpose motion adapter for downstream tasks.

%% file: table/tab_cmp.tex
\begin{table}[t!]
    \centering
    \caption{\textbf{Quantitative comparisons with existing methods.}
    We report PMD ($\times100$) and ELS metrics for humanoid-to-humanoid (H2H) and cross-category transfer (CCT) settings. 
    }
    \vspace{-1em}
\setlength{\tabcolsep}{16pt} %
\resizebox{\linewidth}{!}{%
\begin{tabular}{l|cc|cc}
\toprule
\multirow{2}{*}{\textbf{Methods}} & \multicolumn{2}{c|}{\textbf{H2H}} & \multicolumn{2}{c}{\textbf{CCT}} \\ 
\cmidrule(lr){2-3} \cmidrule(lr){4-5}
 & PMD$\downarrow$ & ELS$\uparrow$ & PMD$\downarrow$ & ELS$\uparrow$ \\
\midrule
NPT~\cite{wang2020neural}          &   6.334   &  0.842    &   9.889   &  0.260    \\
CGT~\cite{chen2022geometry}        &   5.687   &  0.887    &   6.314   &  0.744    \\
SFPT~\cite{liao2022skeleton}       &   3.616   &  0.888    &   4.312   &  0.913    \\
TapMo~\cite{zhang2024tapmo}        &   5.078   &  0.877    &   4.883   &  0.922    \\
\midrule
w/o Eq.~\ref{eq.rotavg} (A1)       &   4.439   &  0.920    &   4.524   &  0.924    \\
w/o Eq.~\ref{eq.priorloss} (A2)    &   4.161   &  0.920    &   4.655   &  0.922    \\
w/o Eq.~\ref{eq.textsim} (A3)      &   4.268   &  0.919    &   4.612   &  0.922    \\
\midrule
{\name} (\textbf{Ours})                     & \textbf{3.570} & \textbf{0.923} & \textbf{4.264} & \textbf{0.927} \\
\bottomrule
\end{tabular}}
\label{tab.cmp}
\vspace{-15pt}
\end{table}

%% file: 5-abl.tex
\section{Ablation Study}

\myparagraph{Settings.}
To assess the contribution of each key component in {\name}, we conduct ablation studies by removing or replacing individual modules while keeping other parts unchanged for fair comparison.  
Specifically, we evaluate:  
\textbf{A1} (w/o Eq.~\ref{eq.rotavg}): replacing our rotation initialization with a simple equal weighted sum instead of the blending scheme in Eq.~\ref{eq.rotavg};  
\textbf{A2} (w/o Eq.~\ref{eq.priorloss}): removing the pose prior regularization; and  
\textbf{A3} (w/o Eq.~\ref{eq.textsim}): pretraining the correspondence module $\mathcal{G}$ using hierarchical correspondence algorithm~\cite{zhan2024charactermixer,xu2022hierarchical} instead of text-based supervision.

\input{figure/fig_t2m}

\input{figure/fig_abl}

\myparagraph{Results.}
The qualitative and quantitative comparisons are shown in Fig.~\ref{fig.abl} and Tab.~\ref{tab.cmp}, respectively.  
Without the pose prior (A2), unnatural deformations such as joint twisting and self-intersections occur, showing that motion priors learned from large-scale motion datasets help constrain plausible transformations.  
Using naive rotation averaging (A1) introduces orientation ambiguity, often causing distorted poses ({\eg}, limbs and wings in target characters).  
Finally, replacing text-based supervision with heuristic correspondences (A3) leads to misaligned mappings--{\eg}, a dog’s hind legs being incorrectly matched to the source’s arms--highlighting the necessity of semantic alignment in correspondence learning.  
Overall, the full {\name} achieves the best transfer quality, validating the complementary effects of each component in enabling robust and semantically consistent pose transfer.

%% file: figure/fig_t2m.tex
\begin{figure}
    \centering
    \begin{overpic}
        [trim=0cm 12cm 0cm 10cm,clip,width=\linewidth,grid=false]{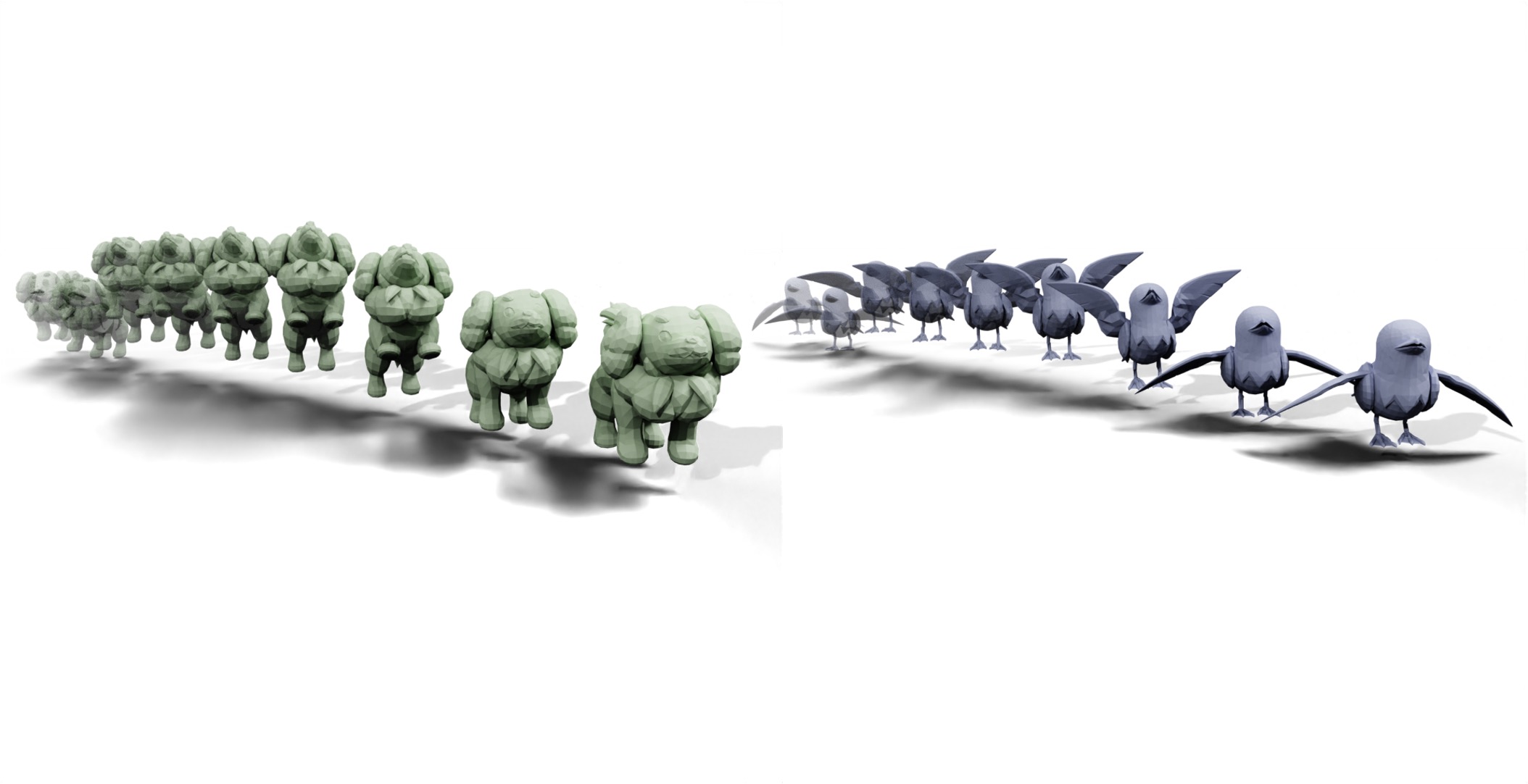}
        \put(28,20){\scriptsize \textbf{Text description: \textit{Jump up and down}}}
    \end{overpic}
    \begin{overpic}
        [trim=0cm 12cm 0cm 10cm,clip,width=\linewidth,grid=false]{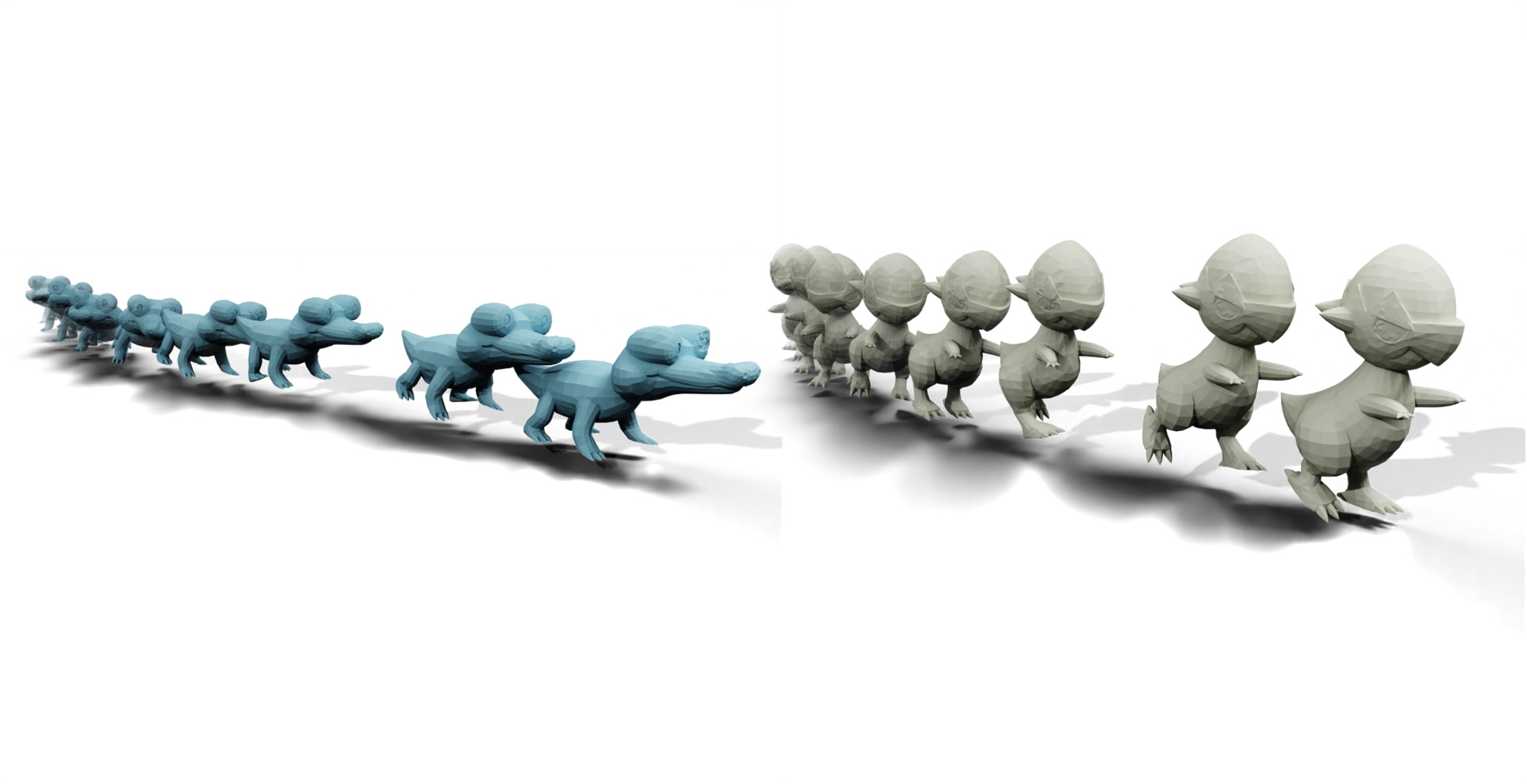}
        \put(25,20.5){\scriptsize \textbf{Text description: \textit{Run and make a slight turn}}}
    \end{overpic}
    \vspace{-2em}
    \caption{\textbf{Application of {\name} on motion generation.}
    We demonstrate that {\name} can be zero-shot integrated with standard text-to-motion models~\cite{jiang2023motiongpt,chen2023executing}, allowing generated human motions to be transferred into diverse target characters.
    }
    \label{fig.t2m}
    \vspace{-5pt}
\end{figure}

%% file: figure/fig_abl.tex
\begin{figure}
    \centering
    \begin{overpic}[trim=0cm 0cm 0cm 0cm,clip,width=\linewidth,grid=false]{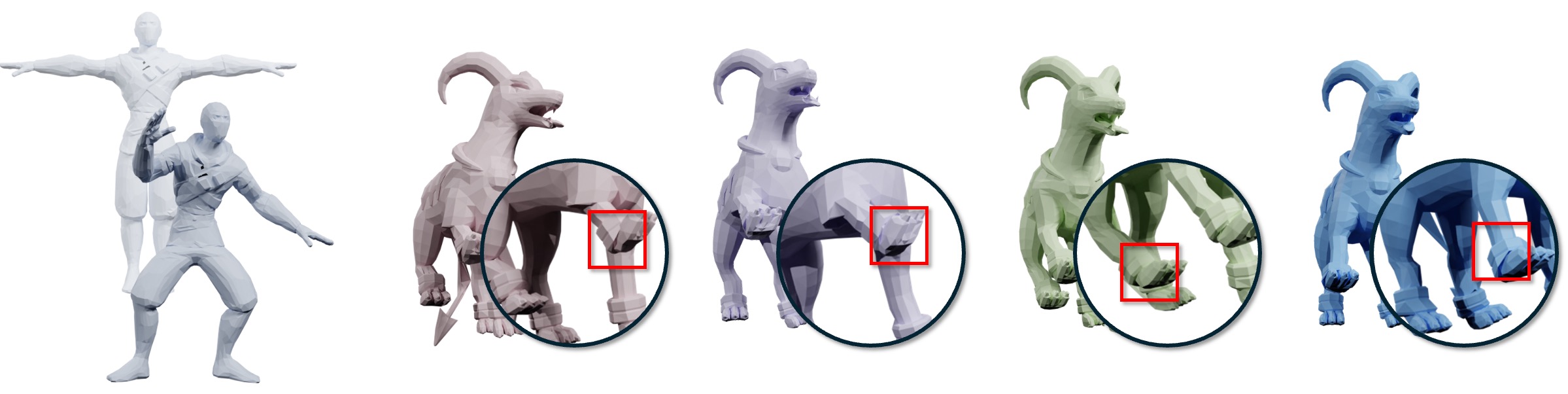}
    \put(3,24){\scriptsize \textbf{Source Pose}}
    \put(24,24){\scriptsize A1 (w/o Eq.~\ref{eq.rotavg})}
    \put(44,24){\scriptsize w/o  A2 (Eq.~\ref{eq.priorloss})}
    \put(64,24){\scriptsize A3 (w/o Eq.~\ref{eq.textsim})}
    \put(86,24){\scriptsize \textbf{\ul{Ours}}}
    \end{overpic}
    \begin{overpic}[trim=0cm 0cm 0cm 0cm,clip,width=\linewidth,grid=false]{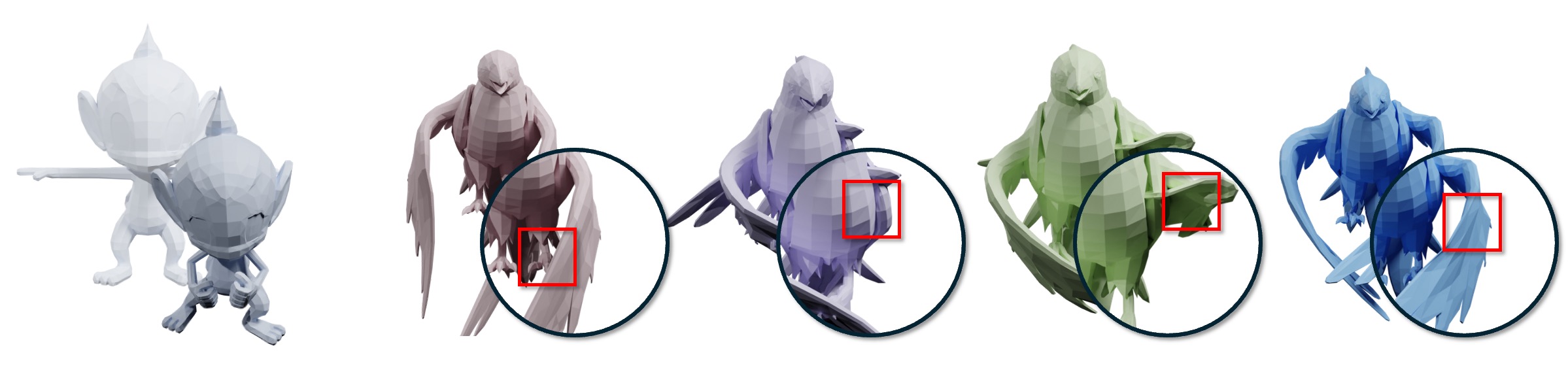}\end{overpic}
    \vspace{-2.5em}
    \caption{
    \textbf{Ablation studies.}
    We qualitatively evaluate the impact of key design choices in {\name}.
    From \textit{left} to \textit{right}: source pose, and the transferred results produced by each ablated variant.
    The results show that each component is essential, and removing any of them noticeably degrades the transfer quality.
    }
    \label{fig.abl}
    \vspace{-15pt}
\end{figure}
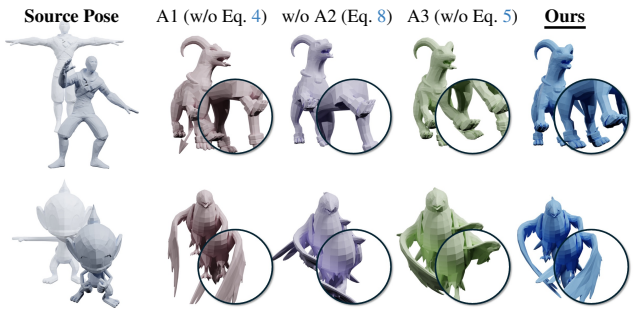

%% file: 6-conc.tex
\section{Conclusion}

This paper introduces {\name}, a cascade-transformer framework supported by a new large-scale motion dataset, {\dataname}, to advance 3D pose transfer toward a general, category-free setting. To the best of our knowledge, this is among the first works to enable pose transfer across structurally diverse characters.
Benefit from the million-scale pose diversity of {\dataname}, {\name} learns flexible soft correspondences and shape-aware deformations, enabling plausible transfer across characters with drastically different geometries.
We further demonstrate the versatility of {\name} in downstream applications such as animation and virtual content creation. We believe {\dataname} will serve as a valuable resource for broader vision tasks and inspire future research within the community.

%% file: 0-supp.tex
\clearpage
\renewcommand{\thesection}{A\arabic{section}}
\renewcommand{\thesubsection}{A\arabic{section}.\arabic{subsection}}
\renewcommand{\thefigure}{A\arabic{figure}}
\renewcommand{\thetable}{A\arabic{table}}
\renewcommand{\theequation}{A\arabic{equation}}

\setcounter{section}{0}
\setcounter{figure}{0}
\setcounter{table}{0}
\setcounter{equation}{0}

\maketitlesupplementary

This supplementary material provides additional technical details and presents further visualized results supporting the main paper.
The content is organized as follows:  
First, the notations used throughout the paper is summarized in Section~\ref{sec.supp_notion}.  
Section~\ref{sec.supp_poseprior} describes the details of our pose prior model.  
Section~\ref{sec.supp_imp_details} presents further implementation details.  
Section~\ref{sec.supp_vis} includes additional qualitative visualizations and a user study to strengthen our results.
Finally, Section~\ref{sec.supp_limitaion} discusses the limitations of our method and outlines potential directions for future work.

\section{Notation Table} \label{sec.supp_notion}

For clarity and ease of reference, the key notations used throughout the paper are summarized in Tab.~\ref{tab.supp_notion}.

\input{table/tab_supp_notion}

\section{Details of Pose Prior Transformer} \label{sec.supp_poseprior}

In this section, we provide technical details of the pose prior model, including how we address the challenges arising from varying keypoint lengths and diverse rotation behaviors across characters, as well as the training objectives.

\subsection{Architecture of Pose Prior Model}

We observe that without proper priors or regularization, the pose transfer model often suffers from severe degeneration such as collapse. However, directly learning unified rotation representations across characters with highly diverse geometries and skeletal structures is non-trivial.  
To address this, we leverage our motion dataset to train a probabilistic prior model that explicitly captures the likelihood of skeletal structures together with their associated rotations.

Formally, the rotation of each keypoint is represented by a quaternion $\mathbf{q}\in \mathbb{S}^3$~\cite{liao2022skeleton}. 
One option is to parameterize this distribution using the Bingham distribution~\cite{gilitschenski2019deep}. However, the Bingham distribution requires strong constraints on its parameters, which limits model expressiveness~\cite{sengupta2021hierarchical}.
Instead, we follow previous works~\cite{yin2022fishermatch,sengupta2021hierarchical,mohlin2020probabilistic} to map quaternions to attitude matrices $\mathbf{R} = A(\mathbf{q}) \in \mathrm{SO}(3)$ and model them using the \emph{matrix-Fisher distribution}, which defines a probability density over $\mathrm{SO}(3)$:
\begin{equation}
    p(\mathbf{R}_k \mid \mathbf{F}_k) 
    = \frac{1}{c(\mathbf{F}_k)} \exp \big( \operatorname{tr}(\mathbf{F}_k^\top \mathbf{R}_k) \big), 
    % = \mathcal{M}(\mathbf{R}_k;\mathbf{F}_k),
    \label{eq.matrixfisher}
\end{equation}
where $\mathbf{F}_k \in \mathbb{R}^{3 \times 3}$ is the distribution parameter of the $k$-th keypoint, and $c(\mathbf{F}_k)$ is the normalization constant.

\input{figure_supp/fig_supp_dataset}

\input{figure_supp/fig_supp_main_res} 

Similar to the cascade-transformer design of {\name}, to capture the joint distribution of rotations across all keypoints of arbitrary characters, we design a transformer-based pose prior model $\mathcal{F}$. It estimates $p(\bar{\mathbf{C}}, \mathbf{C} ;\mathbf{f}_{\bar{\mathbf{V}}},\mathbf{f}_{\mathbf{V}}) = \prod_{k=1}^{K}p(\mathbf{R}_k\mid\mathbf{F}_k)$, where $\bar{\mathbf{C}}$ and $\mathbf{C}$ denote canonical and posed keypoints, and $\mathbf{f}_{\bar{\mathbf{V}}}, \mathbf{f}_{\mathbf{V}}$ are geometry features extracted from the corresponding meshes.

Specifically, we concatenate $\mathbf{f}_{\bar{\mathbf{V}}}$ and $\mathbf{f}_{\mathbf{V}}$ and project them through the shape projector to obtain the shape tokens $f_{\mathbf{M}}\in\mathbb{R}^{l_{\mathcal{E}} \times d_c}$.
For the keypoints tokens, we concatenate the canonical and posed keypoint coordinates $\bar{\mathbf{C}}$ and $\mathbf{C}$, and map them into a $d_c$-dimensional latent representation $f_{\mathbf{C}} \in \mathbb{R}^{K \times d_c}$ via keypoint encoder.
The concatenated tokens $[f_{\mathbf{M}}, f_{\mathbf{C}}]$ are then fed into transformer blocks, which applies attention mechanism to model interactions among keypoints while conditioning on global geometry.
Finally, an MLP decoder maps the latent representations to a set of matrix-Fisher distribution parameters $\{\hat{\mathbf{F}}_1,\cdots, \hat{\mathbf{F}}_{K}\}$, where each $\hat{\mathbf{F}}_k$ models the rotation distribution of the $k$-th keypoint.

\subsection{Training Objective Functions}
Following previous works~\cite{sengupta2021hierarchical}, $\mathcal{F}$ is trained with the negative log-likelihood (NLL) of the ground-truth rotations, with pose sampled from {\dataname} and Mixamo~\cite{mixamo2025}:
\begin{equation}
\resizebox{0.8\columnwidth}{!}{%
$
\mathcal{L}_{\text{NLL}} = \sum\nolimits_{k=1}^{K} \left( \log c(\mathbf{F}_k) - \operatorname{tr}\left( \mathbf{F}_k^\top A(\mathbf{q}_k) \right) \right).
$}
\end{equation}

Additionally, we adopt differentiable rejection sampling~\cite{kent2013new} to draw $n$ candidate rotations from the predicted distributions. Combined with the estimated translation vectors, the sampled characters $\hat{\mathbf{V}}$ are reconstructed via linear blend skinning and supervised against the ground-truth $\mathbf{V}$ using a reconstruction loss:
\begin{equation}
\resizebox{0.95\columnwidth}{!}{%
    $\mathcal{L}_{\text{sample}} 
    = \frac{1}{n} \sum\nolimits_{i=1}^{n} 
    \big\| \hat{\mathbf{V}}^{(i)} - \mathbf{V} \big\|_2^2,
    \; \mathbf{q}^{(i)}_{k} \sim p\left(A(\mathbf{q}_{k}) \mid \mathbf{F}_k\right).
    $}
\end{equation}

The pose prior transformer is trained using AdamW~\cite{loshchilov2018decoupled} with an initial learning rate of $1e{-4}$, a mini-batch size of $256$, for a total of $5$ epochs.
As such, once the probability model is trained, it predicts the distribution parameters for a given canonical–posed keypoint pair. During pose transfer, we regularize the predicted rotations by maximizing their likelihood under these learned distributions, ensuring that the estimated joint rotations remain plausible for the given character geometry.

\section{Implementation Details} \label{sec.supp_imp_details}

\subsection{Model Details of {\name}}

For all modules--the pose prior transformer $\mathcal{F}$, the correspondence transformer $\mathcal{G}$, and the pose transfer transformer $\mathcal{H}$--we adopt a similar architectural design. 
The keypoint encoder and shape projector first map their respective inputs into $256$-dimensional latent tokens, keypoint encoder is implemented with a $2$-layer MLP and shape projector is a linear layer, with hidden dimension of $1{,}024$.

For $\mathcal{F}$, $\mathcal{G}$, and $\mathcal{H}$, each module adopts a $6$-layer stacked transformer encoder, where every layer comprises a multi-head self-attention (MHSA) module (with $8$ heads) followed by a $2$-layer MLP. The MLP uses a hidden dimension of $2{,}048$. 
The distribution decoder of $\mathcal{F}$ is a $2$-layer MLP with a hidden dimension of $128$ and nonlinear activation.
For the correspondence module $\mathcal{G}$, the learnable weights $\mathbf{A}$ is parameterized with a hidden dimension of $256$.
The transformation decoder of $\mathcal{H}$ is implemented as an MLP with a hidden dimension of $256$.

\subsection{Details of Dataset Split}

For the AMASS dataset, we follow the standard protocol in prior works~\cite{liao2022skeleton,zhou2020unsupervised} and split the motions into training and validation sets.
For Mixamo, we use $97$ characters for training and $11$ for testing.
For {\dataname}, we split the dataset into $780$ training characters, $109$ validation characters, and $86$ test characters.
Across these sources, we use a total of $4.21$ million pose samples to train the pose prior transformer $\mathcal{F}$.
For the correspondence transformer $\mathcal{G}$, we construct $384k$ canonical-pose pairs for training.
To train the pose transfer transformer $\mathcal{H}$, we sample $100k$ source-target character pairs from the training split at each epoch, drawing random pose for each pair during every iteration.

\input{figure_supp/fig_supp_cycle_cmp}

\section{Additional Visualized Results} \label{sec.supp_vis}

\subsection{Details of {\dataname}}

As stated in the main paper, {\dataname} provides character-level motion sequences, forming a large-scale corpus of diverse 3D character poses. Each character is associated with a set of predefined motion clips spanning various action categories. On average, a character contains approximately $30$ actions, with the number ranging from $3$ to $102$.
Fig.~\ref{fig.supp_dataset} further illustrates the diversity of poses and character types captured in our dataset.

\subsection{Visualized Results from {\name}}

In Fig.~\ref{fig.supp_mainres}, \ref{fig.supp_mainres2}, and \ref{fig.supp_mainres3}, we present additional qualitative pose transfer results produced by {\name}. For each source character, we transfer $5$ distinct poses to $5$ target characters spanning diverse categories. The results further demonstrate that {\name} can reliably transfer poses across structurally different characters, faithfully preserving geometric details while capturing the pose characteristics from the source characters.

\subsection{Cycle Consistency Comparison}

As a supplement to our quantitative evaluation, we provide additional visual comparisons of pose transfer quality in Fig.~\ref{fig.supp_cycle_cmp}. 
For each method, we show both the transferred target poses and the corresponding cycled-source reconstructions.
The first two rows show examples of humanoid-to-humanoid transfer, where we also include cycle-consistency results for reference.
The remaining three rows illustrate a variety of cross-category transfer cases, covering challenging scenarios with large geometric and topological discrepancies.
These visualizations qualitatively confirm our quantitative findings: {\name} achieves the smallest PMD while preserving mesh smoothness (highest ELS), producing more plausible transfer results than prior baselines.

\subsection{User Study}

In Tab.~\ref{tab.rebuttal_user_study}, we report a user study evaluating human perception of pose transfer quality in comparison with baseline methods. We recruited $50$ participants with diverse technical expertise in computer vision and graphics via \textit{Prolific} to assess $20$ transfer pairs. Participants rated each method on a $1$–$5$ scale in terms of pose similarity and geometric quality, and selected the best-performing method for each pair.
The results show that our method achieves the highest average scores in both pose similarity ($4.076$) and geometric quality ($4.102$), and is chosen as the best method in $60.0\%$ of the votes. These findings are consistent with our quantitative evaluations and overall benchmark rankings.

\input{table/tab_user_study}

\section{Limitation \& Future Work} \label{sec.supp_limitaion}

Although {\name} achieves state-of-the-art performance compared to existing baselines, it still has several limitations.
In this section, we discuss the limitations of our work and outline several future research directions.

First, our framework relies on keypoints and skin weights predicted by pretrained models.
Errors in this stage may propagate to the downstream pose transfer pipeline and negatively affect the final results. In the future, we plan to explore an optimization framework that jointly updates the skin weights, keypoints, and target transformations, such that they coherently contribute to the final transfer quality.

Second, as the exploration of efficient transformer architectures is beyond the scope of this work, {\name} adopts computationally expensive vanilla attention implementations.
An important extension would be to incorporate more efficient attention mechanisms ({\eg}, linear or sparse attention) to reduce computational cost while maintaining--or potentially improving--the quality of pose transfer.

Finally, while we demonstrate that {\name} can serve as a plug-and-play module for zero-shot text-to-any-character motion generation, the current pipeline does not explicitly enforce temporal consistency across frames.
Incorporating temporal modeling could significantly improve motion-level coherence and stability. In future work, we plan to leverage the dataset introduced in this paper to further advance 4D generation and general motion synthesis.

%% file: table/tab_supp_notion.tex
\begin{table}[t!]
\centering
\caption{\textbf{Summary of important notations.}
A consolidated reference of the key variables and symbols used in {\name}, grouped according to the modules introduced in the main paper.}
    \vspace{-1em}
\resizebox{\linewidth}{!}{%
    \begin{tabular}{l|l}
    \toprule
    \textbf{Notation} & \textbf{Description} \\ \midrule
    $\mathbf{\bar{V}}^{\text{src}}$ & vertices of source character in canonical pose \\
    $\mathbf{V}^{\text{src}}$ & vertices of posed source character \\
    $N^{\text{src}}$ & number of vertices in source character \\
    $\mathbf{\bar{V}}^{\text{tgt}}$ &  vertices of target character in canonical pose  \\
    $\hat{\mathbf{V}}^{\text{tgt}}$ & vertices of posed target character (predicted) \\
    $N^{\text{tgt}}$ & number of vertices in target character \\
    $\bar{\mathbf{C}}^{\text{src}}$  & canonical keypoints of source character \\
    $\bar{\mathbf{C}}^{\text{tgt}}$  & canonical keypoints of target character \\
    $\nkptsrc$ & number of keypoint of source character  \\
    $\nkpttgt$ & number of keypoint of target character  \\
    $f(\mathbf{V}^{\text{A}}, \mathbf{\bar{V}}^{\text{A}}, \mathbf{\bar{V}}^{\text{B}}) \to 
    \mathbf{\hat{V}}^{\text{B}}$ &  transferring pose from character A to B \\ \midrule
    $ \mathcal{E}$ & pretrained shape encoder \\
    $\mathbf{f}_{\bar{\mathbf{V}}^{\text{src}}}$ & shape feature of canonical source character\\
    $\mathbf{f}_{{\mathbf{V}}^{\text{src}}}$ & shape feature of posed source character\\
    $\mathbf{f}_{\bar{\mathbf{V}}^{\text{tgt}}}$ &  shape feature of canonical target character\\
    $\delta_\mathbf{f}$ & residual shape feature of source character \\ \midrule
    $\mathcal{F}$ &  pose prior transformer \\
    $A(\mathbf{q})$ & attitude matrix of quaternion $\mathbf{q}$ \\
    $f_{\mathbf{M}}$ & shape tokens for $\mathcal{F}$  \\
    $f_{\mathbf{C}}$ & keypoint tokens for $\mathcal{F}$ \\
    $\hat{\mathbf{F}}$ & parameters of matrix-Fisher distribution \\
    $c(\mathbf{F}_k)$ & distribution normalization constant {\wrt} $\mathbf{F}_k$ \\ 
    \midrule
    $\mathcal{G}$ & correspondence transformer  \\
    $g_{\mathbf{M}}$ & shape tokens for $\mathcal{G}$  \\
    $g_{\mathbf{C}}$ & keypoint tokens for $\mathcal{G}$  \\
    $\mathbf{g}^{\text{src}}$ & shape-aware latent representations of source keypoints\\
    $\mathbf{g}^{\text{tgt}}$ & shape-aware latent representations of target keypoints\\
    $\mathbf{A}$ & learnable weight for affinity matrix of $\mathcal{G}$ \\
    $\mathbf{S}$ & similarity matrix of source and target keypoints \\
    $\mathbf{f}_k$ & CLIP latent feature of keypoint $\mathbf{c}_k$  \\
    $\mathbf{S}_{\cos}$ & CLIP-based cosine similarity of keypoint pairs \\
    $\mathbf{M}$ & doubly stochastic correspondence matrix \\ \midrule
    $\mathcal{H}$ & pose transfer transformer \\
    $h_{\mathbf{M}}$ & shape tokens for $\mathcal{H}$ \\
    $h_{\mathbf{C}}$ & keypoint tokens for $\mathcal{H}$ \\
    $\mathbf{T}^{\text{src}} = \{\mathbf{T}^{\text{src}}_1, \cdots, \mathbf{T}^{\text{src}}_{\nkptsrc}\}$             & per-keypoint transformations for source character\\
    $\mathbf{T}^{\text{tgt}} = \{\hat{\mathbf{T}}^{\text{tgt}}_1, \ldots, \hat{\mathbf{T}}^{\text{tgt}}_{\nkpttgt}\}$ & per-keypoint transformations for target character\\
    $\mathbf{q}_i$ & rotation quaternion of $i$-th keypoint \\
    $\mathbf{t}_i$ & translation vector of $i$-th keypoint \\
    \bottomrule
    \end{tabular}}
    \vspace{-15pt}
\label{tab.supp_notion}
\end{table}

%% file: figure_supp/fig_supp_dataset.tex
\begin{figure*}[t!]
    \centering
    \begin{overpic}[trim=9cm 1cm 10cm 17cm,clip,width=\linewidth,grid=false]{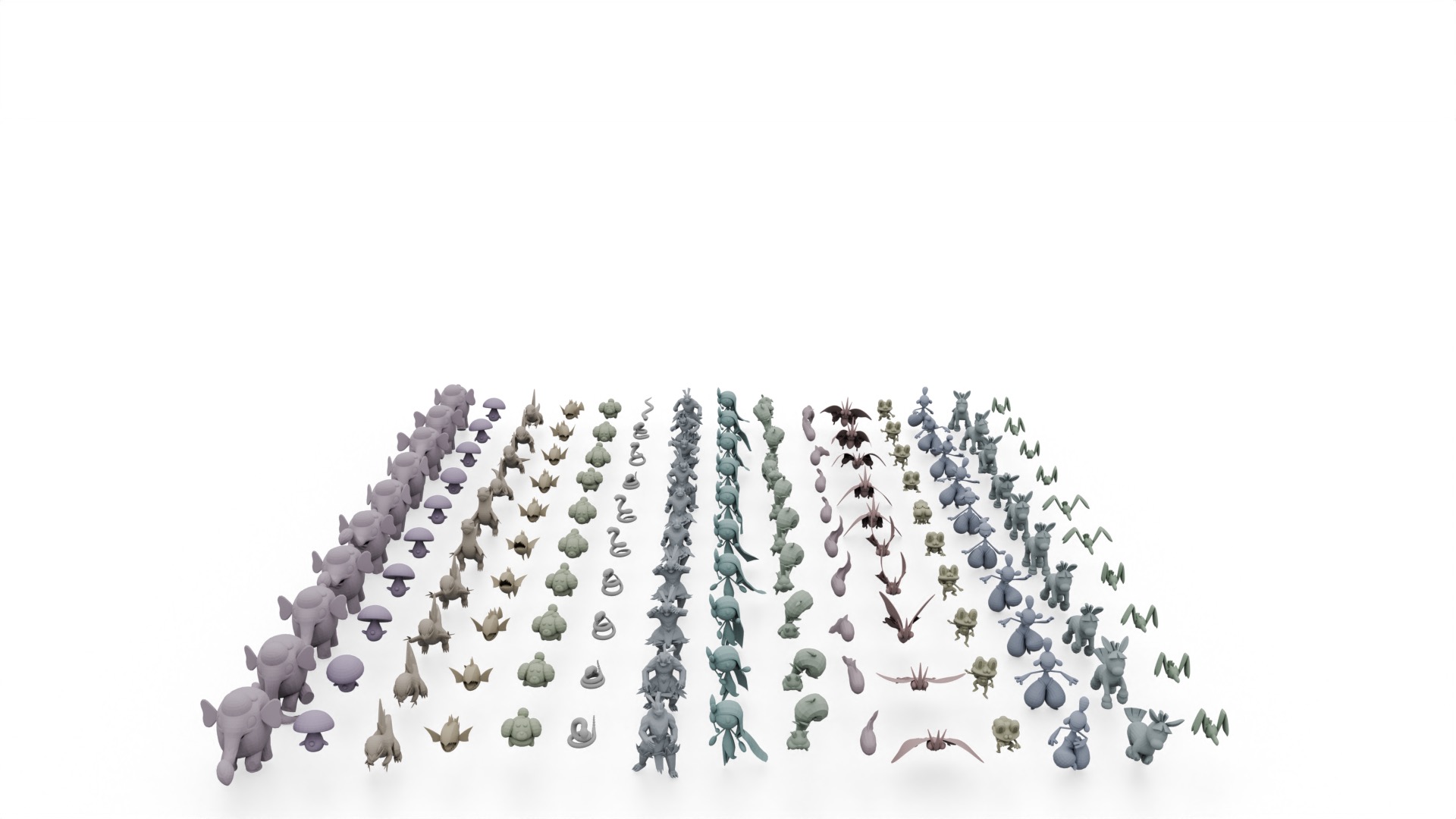}\end{overpic}
    \vspace{-2em}
    \caption{
    \textbf{Additional pose examples from {\dataname}.} {\dataname} contains diverse and high-quality character poses covering a wide spectrum of species and morphological structures. From \textit{left} to \textit{right}, we present representative pose samples across characters.
    }
    \label{fig.supp_dataset}
    % \vspace{-10pt}
\end{figure*}

%% file: figure_supp/fig_supp_main_res.tex
\begin{figure*}[ht!]
    \centering
    \begin{overpic}[trim=0cm 0cm 0cm 0cm,clip,width=\linewidth,grid=false]{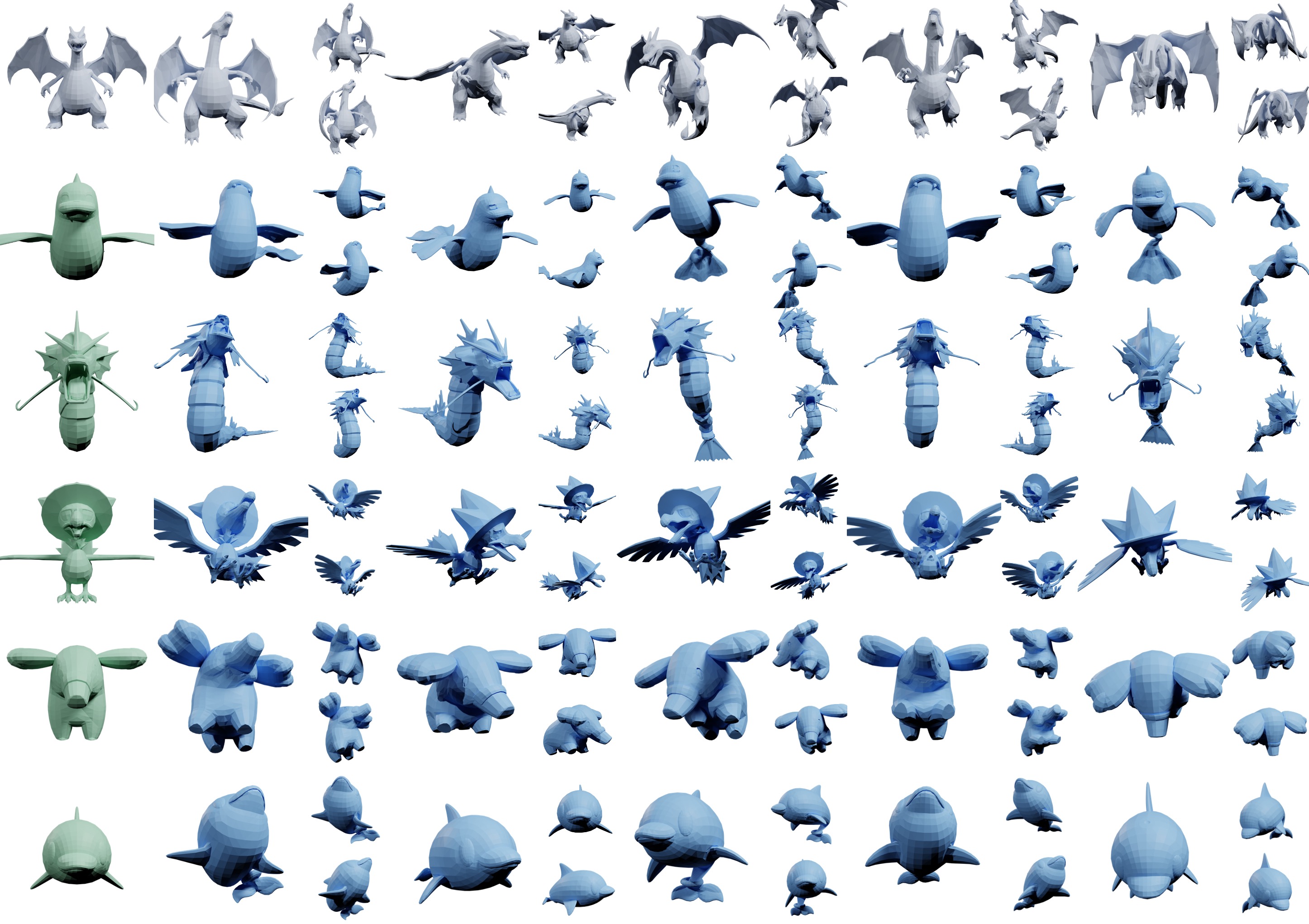}
    \put(4,70){\scriptsize \textbf{Source}}
    \put(3.5,58){\scriptsize \textbf{Target 1}}
    \put(3.5,47){\scriptsize \textbf{Target 2}}
    \put(3.5,34){\scriptsize \textbf{Target 3}}
    \put(3.5,22){\scriptsize \textbf{Target 4}}
    \put(3.5,10){\scriptsize \textbf{Target 5}}
    \put(15,70){\scriptsize \textbf{Pose 1}}
    \put(34,70){\scriptsize \textbf{Pose 2}}
    \put(50,70){\scriptsize \textbf{Pose 3}}
    \put(69,70){\scriptsize \textbf{Pose 4}}
    \put(86,70){\scriptsize \textbf{Pose 5}}
    \end{overpic}
    \vspace{-2em}
    \caption{
    \textbf{Qualitative results of {\name} (part I).} 
    We present pose transfer results across a wide range of character categories, with each example rendered from three viewpoints. 
    From \textit{left} to \textit{right}: the canonical character followed by its transferred results under five different poses. 
    The \textit{1st} row shows the source character and the five input poses; the \textit{2nd}–\textit{6th} rows show the corresponding transferred poses for each target character.
    }
    \label{fig.supp_mainres}
    \vspace{-10pt}
\end{figure*}

\begin{figure*}[ht!]
    \centering
    \begin{overpic}[trim=0cm 0cm 0cm 0cm,clip,width=\linewidth,grid=false]{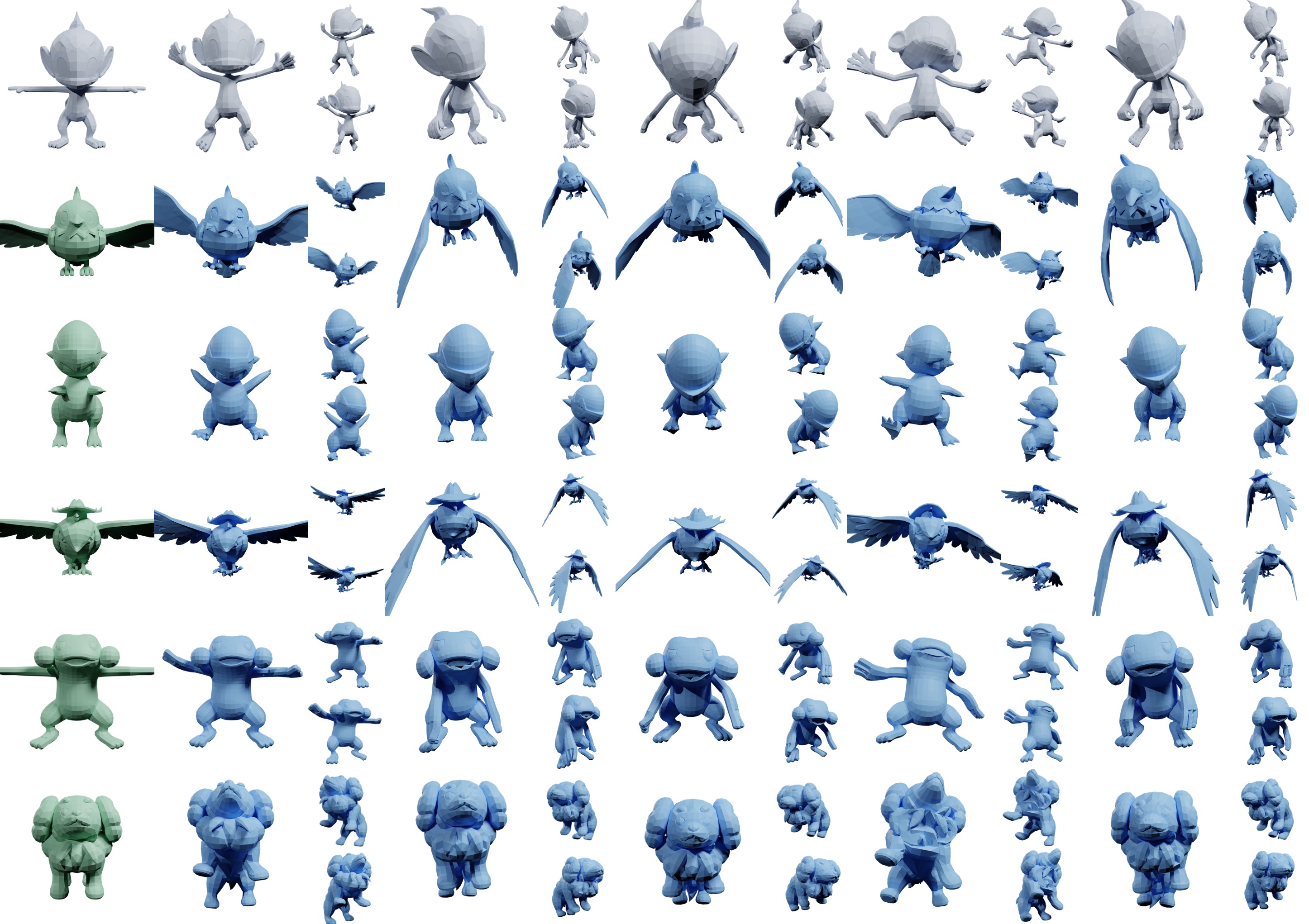}
    \put(4,71){\scriptsize \textbf{Source}}
    \put(3.5,57){\scriptsize \textbf{Target 1}}
    \put(3.5,47){\scriptsize \textbf{Target 2}}
    \put(3.5,34){\scriptsize \textbf{Target 3}}
    \put(3.5,23){\scriptsize \textbf{Target 4}}
    \put(3.5,11){\scriptsize \textbf{Target 5}}
    \put(15,71){\scriptsize \textbf{Pose 1}}
    \put(33,71){\scriptsize \textbf{Pose 2}}
    \put(51,71){\scriptsize \textbf{Pose 3}}
    \put(69,71){\scriptsize \textbf{Pose 4}}
    \put(86,71){\scriptsize \textbf{Pose 5}}
    \end{overpic}
    \vspace{-2em}
    \caption{
    \textbf{Qualitative results of {\name} (part II).} 
    We present pose transfer results across a wide range of character categories, with each example rendered from three viewpoints. 
    From \textit{left} to \textit{right}: the canonical character followed by its transferred results under five different poses. 
    The \textit{1st} row shows the source character and the five input poses; the \textit{2nd}–\textit{6th} rows show the corresponding transferred poses for each target character.
    }
    \label{fig.supp_mainres2}
    \vspace{-15pt}
\end{figure*}

\begin{figure*}[ht!]
    \centering
    \begin{overpic}[trim=0cm 0cm 0cm 0cm,clip,width=\linewidth,grid=false]{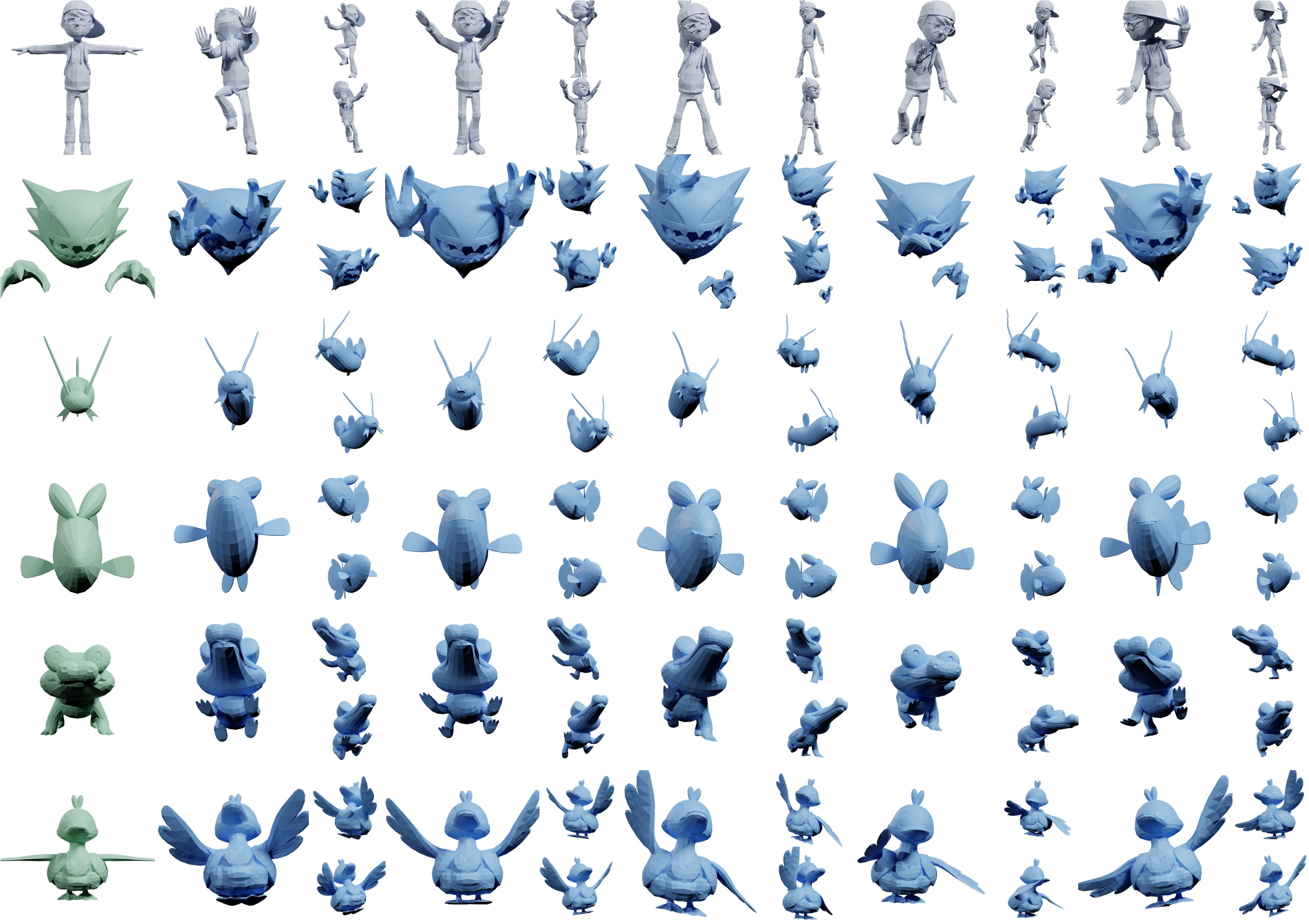}
    \put(4,71){\scriptsize \textbf{Source}}
    \put(3.5,57){\scriptsize \textbf{Target 1}}
    \put(3.5,45){\scriptsize \textbf{Target 2}}
    \put(3.5,35){\scriptsize \textbf{Target 3}}
    \put(3.5,22){\scriptsize \textbf{Target 4}}
    \put(3.5,11){\scriptsize \textbf{Target 5}}
    \put(15,71){\scriptsize \textbf{Pose 1}}
    \put(33,71){\scriptsize \textbf{Pose 2}}
    \put(51,71){\scriptsize \textbf{Pose 3}}
    \put(69,71){\scriptsize \textbf{Pose 4}}
    \put(86,71){\scriptsize \textbf{Pose 5}}
    \end{overpic}
    \vspace{-2em}
    \caption{
    \textbf{Qualitative results of {\name} (part III).} 
    We present pose transfer results across a wide range of character categories, with each example rendered from three viewpoints. 
    From \textit{left} to \textit{right}: the canonical character followed by its transferred results under five different poses. 
    The \textit{1st} row shows the source character and the five input poses; the \textit{2nd}–\textit{6th} rows show the corresponding transferred poses for each target character.
    }
    \label{fig.supp_mainres3}
    \vspace{-15pt}
\end{figure*}

%% file: figure_supp/fig_supp_cycle_cmp.tex
\begin{figure*}[ht!]
    \centering
    \begin{overpic}[trim=0cm -3cm 0cm 0cm,clip,width=\linewidth,grid=false]{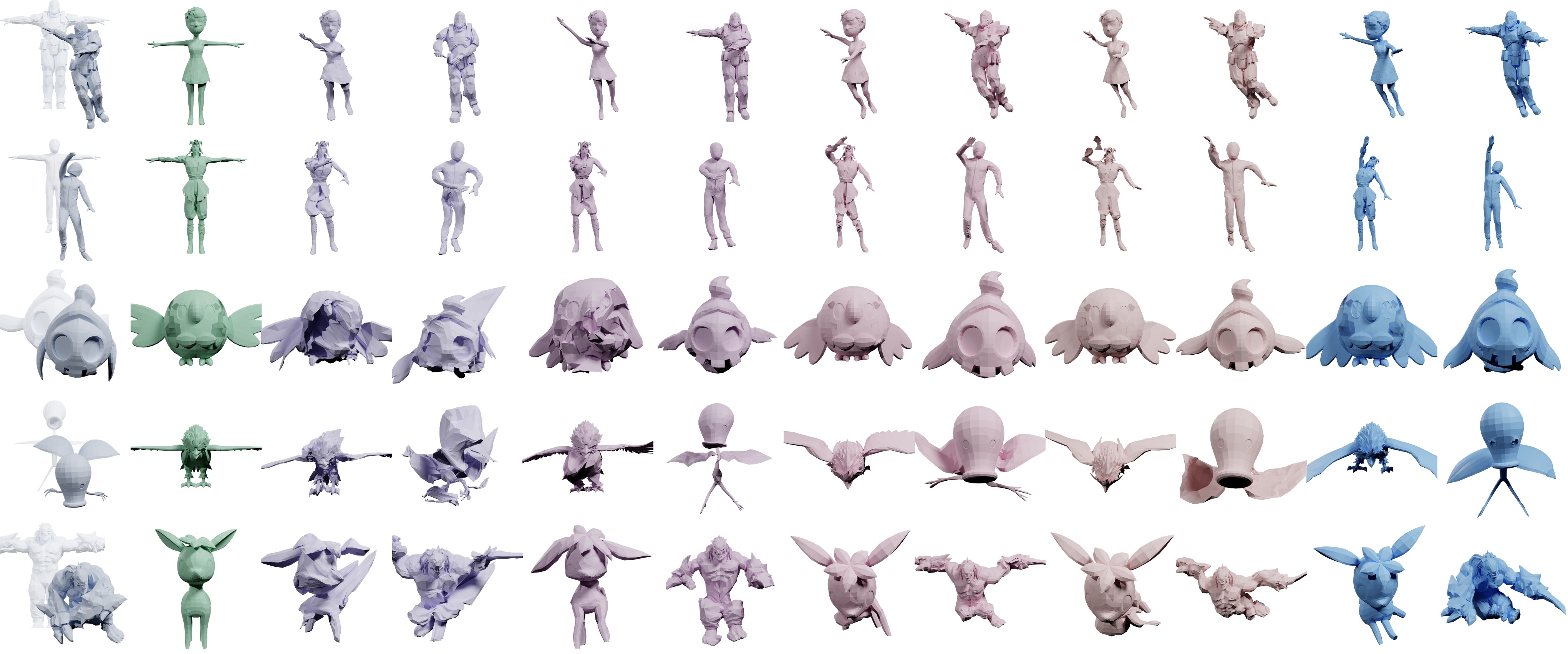}
    \put(1,44){\scriptsize \textbf{Source Pose}}
    \put(11,44){\scriptsize \textbf{\ul{Target}}}
    \put(23,44){\scriptsize \mbox{NPT~\cite{wang2020neural}}}
    \put(39,44){\scriptsize \mbox{CGT~\cite{chen2022geometry}}}
    \put(56,44){\scriptsize \mbox{SFPT~\cite{liao2022skeleton}}}
    \put(72,44){\scriptsize \mbox{TapMo~\cite{zhang2024tapmo}}}
    \put(90,44){\scriptsize \textbf{\ul{Ours}}}
    \put(85,1){\scriptsize \textbf{{src$\rightarrow$tgt}}}
    \put(93,1){\scriptsize \textbf{{tgt$\rightarrow$src}}}

    \put(68,1){\scriptsize \textbf{{src$\rightarrow$tgt}}}
    \put(77,1){\scriptsize \textbf{{tgt$\rightarrow$src}}}

    \put(52,1){\scriptsize \textbf{{src$\rightarrow$tgt}}}
    \put(60,1){\scriptsize \textbf{{tgt$\rightarrow$src}}}

    \put(35,1){\scriptsize \textbf{{src$\rightarrow$tgt}}}
    \put(44,1){\scriptsize \textbf{{tgt$\rightarrow$src}}}

    \put(18,1){\scriptsize \textbf{{src$\rightarrow$tgt}}}
    \put(26,1){\scriptsize \textbf{{tgt$\rightarrow$src}}}
    \end{overpic}
    \vspace{-2em}
    \caption{
    \textbf{Qualitative cycle-consistency comparisons with existing methods.}
    From \textit{left} to \textit{right}: source character, target character, and bidirectional pose transfer results (source$\rightarrow$target and target$\rightarrow$source) produced by different methods.
    {\name} consistently yields higher-quality transfers with more realistic poses and fewer distortions.
    }
    \label{fig.supp_cycle_cmp}
    \vspace{-15pt}
\end{figure*}

%% file: table/tab_user_study.tex
\begin{table}[t!]
    \centering
    \caption{\textbf{Perceptual user study comparisons with existing methods.} We ask participants to assess samples across two primary dimensions: pose similarity and geometric quality.}
    \vspace{-1em}
    \setlength{\tabcolsep}{4pt}
    \resizebox{\linewidth}{!}{%
    \begin{tabular}{l|ccccc}
        \toprule
        {Rating (1--5)} & NPT~\cite{wang2020neural} & CGT~\cite{chen2022geometry} & SFPT~\cite{liao2022skeleton} &TapMO~\cite{zhang2024tapmo} & \textbf{Ours} \\
        \midrule
        Pose Similarity$\uparrow$  & 1.884 & 2.310 &  3.364 & 3.292 & \textbf{4.076} \\
        Geo. Quality$\uparrow$    & 1.556 & 2.516 & 3.531 & 3.605 & \textbf{4.102} \\
        \bottomrule
    \end{tabular}}
    \label{tab.rebuttal_user_study}
    \vspace{-15pt}
\end{table}

%% file: main.bbl
\begin{thebibliography}{83}
\providecommand{\natexlab}[1]{#1}
\providecommand{\url}[1]{\texttt{#1}}
\expandafter\ifx\csname urlstyle\endcsname\relax
  \providecommand{\doi}[1]{doi: #1}\else
  \providecommand{\doi}{doi: \begingroup \urlstyle{rm}\Url}\fi

\bibitem[mix(2025)]{mixamo2025}
Mixamo.
\newblock Online service by Adobe., 2025.
\newblock Accessed: Jan 2025.

\bibitem[Aberman et~al.(2020)Aberman, Li, Lischinski, Sorkine-Hornung, Cohen-Or, and Chen]{aberman2020skeleton}
Kfir Aberman, Peizhuo Li, Dani Lischinski, Olga Sorkine-Hornung, Daniel Cohen-Or, and Baoquan Chen.
\newblock {Skeleton-Aware Networks for Deep Motion Retargeting}.
\newblock \emph{ACM Transactions on Graphics}, 39\penalty0 (4):\penalty0 62:1--62:14, 2020.

\bibitem[Aigerman et~al.(2022)Aigerman, Gupta, Kim, Chaudhuri, Saito, and Groueix]{aigerman2022neural}
Noam Aigerman, Kunal Gupta, Vladimir~G Kim, Siddhartha Chaudhuri, Jun Saito, and Thibault Groueix.
\newblock {Neural Jacobian Fields: Learning Intrinsic Mappings of Arbitrary Meshes}.
\newblock \emph{ACM Transactions on Graphics}, 41\penalty0 (4):\penalty0 109:1--109:17, 2022.

\bibitem[Baran et~al.(2009)Baran, Vlasic, Grinspun, and Popovi{\'c}]{baran2009semantic}
Ilya Baran, Daniel Vlasic, Eitan Grinspun, and Jovan Popovi{\'c}.
\newblock Semantic deformation transfer.
\newblock \emph{ACM Transactions on Graphics}, 28\penalty0 (3):\penalty0 36:1--36:6, 2009.

\bibitem[Ben-Chen et~al.(2009)Ben-Chen, Weber, and Gotsman]{ben2009spatial}
Mirela Ben-Chen, Ofir Weber, and Craig Gotsman.
\newblock Spatial deformation transfer.
\newblock In \emph{Proceedings of the 2009 ACM SIGGRAPH/Eurographics Symposium on Computer Animation}, pages 67--74, 2009.

\bibitem[Besl and McKay(1992)]{besl1992method}
Paul~J Besl and Neil~D McKay.
\newblock {Method for registration of 3-D shapes}.
\newblock In \emph{Sensor fusion IV: control paradigms and data structures}, pages 586--606. SPIE, 1992.

\bibitem[Cha et~al.(2025)Cha, Yoon, Seo, and Noh]{cha2025neural}
Sihun Cha, Serin Yoon, Kwanggyoon Seo, and Junyong Noh.
\newblock Neural face skinning for mesh-agnostic facial expression cloning.
\newblock \emph{Computer Graphics Forum}, 44\penalty0 (2), 2025.

\bibitem[Chen et~al.(2021)Chen, Tang, Shi, Peng, Sebe, and Zhao]{chen2021intrinsic}
Haoyu Chen, Hao Tang, Henglin Shi, Wei Peng, Nicu Sebe, and Guoying Zhao.
\newblock {Intrinsic-Extrinsic Preserved GANs for Unsupervised 3D Pose Transfer}.
\newblock In \emph{IEEE/CVF International Conference on Computer Vision}, pages 8630--8639, 2021.

\bibitem[Chen et~al.(2022)Chen, Tang, Yu, Sebe, and Zhao]{chen2022geometry}
Haoyu Chen, Hao Tang, Zitong Yu, Nicu Sebe, and Guoying Zhao.
\newblock {Geometry-contrastive transformer for generalized 3D pose transfer}.
\newblock In \emph{Proceedings of the AAAI Conference on Artificial Intelligence}, pages 258--266, 2022.

\bibitem[Chen et~al.(2023{\natexlab{a}})Chen, Tang, Timofte, Gool, and Zhao]{chen2023lart}
Haoyu Chen, Hao Tang, Radu Timofte, Luc~V Gool, and Guoying Zhao.
\newblock {LART: Neural Correspondence Learning with Latent Regularization Transformer for 3D Motion Transfer}.
\newblock \emph{Advances in Neural Information Processing Systems}, 36:\penalty0 43742--43753, 2023{\natexlab{a}}.

\bibitem[Chen et~al.(2024)Chen, Tang, Adeli, and Zhao]{chen2024towards}
Haoyu Chen, Hao Tang, Ehsan Adeli, and Guoying Zhao.
\newblock {Towards robust 3D pose transfer with adversarial learning}.
\newblock In \emph{IEEE/CVF Conference on Computer Vision and Pattern Recognition}, pages 2295--2304, 2024.

\bibitem[Chen et~al.(2023{\natexlab{b}})Chen, Li, and Lee]{chen2023weakly}
Jinnan Chen, Chen Li, and Gim~Hee Lee.
\newblock {Weakly-supervised 3D pose transfer with keypoints}.
\newblock In \emph{IEEE/CVF International Conference on Computer Vision}, pages 15110--15119, 2023{\natexlab{b}}.

\bibitem[Chen et~al.(2025)Chen, Zhang, Yin, Dou, Chen, Wang, Komura, and Zhang]{chen2025motion2motion}
Ling-Hao Chen, Yuhong Zhang, Zixin Yin, Zhiyang Dou, Xin Chen, Jingbo Wang, Taku Komura, and Lei Zhang.
\newblock {Motion2Motion: Cross-topology Motion Transfer with Sparse Correspondence}.
\newblock In \emph{SIGGRAPH Asia}, pages 150:1--150:11, 2025.

\bibitem[Chen et~al.(2023{\natexlab{c}})Chen, Jiang, Liu, Huang, Fu, Chen, and Yu]{chen2023executing}
Xin Chen, Biao Jiang, Wen Liu, Zilong Huang, Bin Fu, Tao Chen, and Gang Yu.
\newblock Executing your commands via motion diffusion in latent space.
\newblock In \emph{IEEE/CVF conference on Computer Vision and Pattern Recognition}, pages 18000--18010, 2023{\natexlab{c}}.

\bibitem[Deitke et~al.(2023)Deitke, Liu, Wallingford, Ngo, Michel, Kusupati, Fan, Laforte, Voleti, Gadre, et~al.]{deitke2023objaverse}
Matt Deitke, Ruoshi Liu, Matthew Wallingford, Huong Ngo, Oscar Michel, Aditya Kusupati, Alan Fan, Christian Laforte, Vikram Voleti, Samir~Yitzhak Gadre, et~al.
\newblock {Objaverse-XL: A Universe of 10M+ 3D Objects}.
\newblock \emph{Advances in Neural Information Processing Systems}, 36:\penalty0 35799--35813, 2023.

\bibitem[Deng et~al.(2025)Deng, Zhang, Geng, Wu, and Wu]{deng2025anymate}
Yufan Deng, Yuhao Zhang, Chen Geng, Shangzhe Wu, and Jiajun Wu.
\newblock {Anymate: A Dataset and Baselines for Learning 3D Object Rigging}.
\newblock In \emph{SIGGRAPH Conference Papers}, pages 112:1--112:10, 2025.

\bibitem[Du et~al.(2025)Du, Hu, Li, Xu, Huang, Fu, and Liu]{du2025emergent}
Keyu Du, Jingyu Hu, Haipeng Li, Hao Xu, Haibing Huang, Chi{-}Wing Fu, and Shuaicheng Liu.
\newblock {Hierarchical Neural Semantic Representation for 3D Semantic Correspondence}.
\newblock In \emph{SIGGRAPH Asia}, pages 142:1--142:11, 2025.

\bibitem[Gao et~al.(2018)Gao, Yang, Qiao, Lai, Rosin, Xu, and Xia]{gao2018automatic}
Lin Gao, Jie Yang, Yi-Ling Qiao, Yu-Kun Lai, Paul~L Rosin, Weiwei Xu, and Shihong Xia.
\newblock Automatic unpaired shape deformation transfer.
\newblock \emph{ACM Transactions on Graphics}, 37\penalty0 (6):\penalty0 237:1--237:15, 2018.

\bibitem[Gat et~al.(2025)Gat, Raab, Tevet, Reshef, Bermano, and Cohen-Or]{gat2025anytop}
Inbar Gat, Sigal Raab, Guy Tevet, Yuval Reshef, Amit~Haim Bermano, and Daniel Cohen-Or.
\newblock {AnyTop: Character Animation Diffusion with Any Topology}.
\newblock In \emph{SIGGRAPH Conference Papers}, pages 13:1--13:10, 2025.

\bibitem[Gilitschenski et~al.(2020)Gilitschenski, Sahoo, Schwarting, Amini, Karaman, and Rus]{gilitschenski2019deep}
Igor Gilitschenski, Roshni Sahoo, Wilko Schwarting, Alexander Amini, Sertac Karaman, and Daniela Rus.
\newblock {Deep Orientation Uncertainty Learning based on a Bingham Loss}.
\newblock In \emph{International Conference on Learning Representations}, 2020.

\bibitem[Gleicher(1998)]{Gleicher_1998}
Michael Gleicher.
\newblock Retargetting motion to new characters.
\newblock In \emph{SIGGRAPH}, pages 33--42, 1998.

\bibitem[Guo et~al.(2022)Guo, Zou, Zuo, Wang, Ji, Li, and Cheng]{guo2022generating}
Chuan Guo, Shihao Zou, Xinxin Zuo, Sen Wang, Wei Ji, Xingyu Li, and Li Cheng.
\newblock {Generating Diverse and Natural 3D Human Motions from Text}.
\newblock In \emph{IEEE/CVF Conference on Computer Vision and Pattern Recognition}, pages 5152--5161, 2022.

\bibitem[Hong et~al.(2025)Hong, Choi, Kim, Cha, and Noh]{hong2025asmr}
Seokhyeon Hong, Soojin Choi, Chaelin Kim, Sihun Cha, and Junyong Noh.
\newblock {ASMR: Adaptive Skeleton-Mesh Rigging and Skinning via 2D Generative Prior}.
\newblock \emph{Computer Graphics Forum}, 44\penalty0 (2), 2025.

\bibitem[Jiang et~al.(2023)Jiang, Chen, Liu, Yu, Yu, and Chen]{jiang2023motiongpt}
Biao Jiang, Xin Chen, Wen Liu, Jingyi Yu, Gang Yu, and Tao Chen.
\newblock {MotionGPT: Human Motion as a Foreign Language}.
\newblock \emph{Advances in Neural Information Processing Systems}, 36:\penalty0 20067--20079, 2023.

\bibitem[Kavan(2014)]{kavan2014part}
Ladislav Kavan.
\newblock {Part I: Direct Skinning Methods and Deformation Primitives}.
\newblock In \emph{SIGGRAPH Course 2014}, pages 1--11, 2014.

\bibitem[Kent et~al.(2013)Kent, Ganeiber, and Mardia]{kent2013new}
John~T Kent, Asaad~M Ganeiber, and Kanti~V Mardia.
\newblock {A new method to simulate the Bingham and related distributions in directional data analysis with applications}.
\newblock \emph{arXiv preprint arXiv:1310.8110}, 2013.

\bibitem[Kuhn(1955)]{kuhn1955hungarian}
Harold~W Kuhn.
\newblock {The Hungarian method for the assignment problem}.
\newblock \emph{Naval Research Logistics Quarterly}, 2\penalty0 (1-2):\penalty0 83--97, 1955.

\bibitem[Lee et~al.(2025)Lee, Jeong, Moon, Kim, Kim, Kim, and Lee]{leemove}
Wonkwang Lee, Jongwon Jeong, Taehong Moon, Hyeon-Jong Kim, Jaehyeon Kim, Gunhee Kim, and Byeong-Uk Lee.
\newblock {How to Move Your Dragon: Text-to-Motion Synthesis for Large-Vocabulary Objects}.
\newblock In \emph{International Conference on Machine Learning}, 2025.

\bibitem[Li et~al.(2021{\natexlab{a}})Li, Aberman, Hanocka, Liu, Sorkine-Hornung, and Chen]{li2021learning}
Peizhuo Li, Kfir Aberman, Rana Hanocka, Libin Liu, Olga Sorkine-Hornung, and Baoquan Chen.
\newblock Learning skeletal articulations with neural blend shapes.
\newblock \emph{ACM Transactions on Graphics}, 40\penalty0 (4):\penalty0 130:1--130:15, 2021{\natexlab{a}}.

\bibitem[Li et~al.(2023)Li, Won, Clegg, Kim, Rai, and Ha]{li2023ace}
Tianyu Li, Jungdam Won, Alexander Clegg, Jeonghwan Kim, Akshara Rai, and Sehoon Ha.
\newblock {ACE: Adversarial Correspondence Embedding for Cross Morphology Motion Retargeting from Human to Nonhuman Characters}.
\newblock In \emph{SIGGRAPH Asia}, pages 46:1--46:11, 2023.

\bibitem[Li et~al.(2021{\natexlab{b}})Li, Takehara, Taketomi, Zheng, and Nie{\ss}ner]{li20214dcomplete}
Yang Li, Hikari Takehara, Takafumi Taketomi, Bo Zheng, and Matthias Nie{\ss}ner.
\newblock {4DComplete: Non-Rigid Motion Estimation Beyond the Observable Surface}.
\newblock In \emph{IEEE/CVF International Conference on Computer Vision}, pages 12706--12716, 2021{\natexlab{b}}.

\bibitem[Liao et~al.(2022)Liao, Yang, Saito, Pons-Moll, and Zhou]{liao2022skeleton}
Zhouyingcheng Liao, Jimei Yang, Jun Saito, Gerard Pons-Moll, and Yang Zhou.
\newblock {Skeleton-Free Pose Transfer for Stylized 3D Characters}.
\newblock In \emph{European Conference on Computer Vision}, pages 640--656, 2022.

\bibitem[Liu et~al.(2025)Liu, Xu, Yifan, Tan, Xu, Wang, Su, and Shi]{liu2025riganything}
Isabella Liu, Zhan Xu, Wang Yifan, Hao Tan, Zexiang Xu, Xiaolong Wang, Hao Su, and Zifan Shi.
\newblock {RigAnything: Template-Free Autoregressive Rigging for Diverse 3D Assets}.
\newblock \emph{ACM Transactions on Graphics}, 44\penalty0 (4):\penalty0 122:1--122:12, 2025.

\bibitem[Loper et~al.(2015)Loper, Mahmood, Romero, Pons-Moll, and Black]{loper2015smpl}
Matthew Loper, Naureen Mahmood, Javier Romero, Gerard Pons-Moll, and Michael~J Black.
\newblock {SMPL: a skinned multi-person linear model}.
\newblock \emph{ACM Transactions on Graphics}, 34\penalty0 (6):\penalty0 248:1--248:16, 2015.

\bibitem[Loshchilov and Hutter(2019)]{loshchilov2018decoupled}
Ilya Loshchilov and Frank Hutter.
\newblock {Decoupled Weight Decay Regularization}.
\newblock In \emph{International Conference on Learning Representations}, 2019.

\bibitem[Luo et~al.(2023)Luo, Cai, Dong, Ming, Qiu, Zhan, and Han]{luo2023rabit}
Zhongjin Luo, Shengcai Cai, Jinguo Dong, Ruibo Ming, Liangdong Qiu, Xiaohang Zhan, and Xiaoguang Han.
\newblock {RaBit: Parametric Modeling of 3D Biped Cartoon Characters with a Topological-Consistent Dataset}.
\newblock In \emph{Proceedings of the IEEE/CVF Conference on Computer Vision and Pattern Recognition}, pages 12825--12835, 2023.

\bibitem[Magnet et~al.(2022)Magnet, Ren, Sorkine-Hornung, and Ovsjanikov]{magnet2022smooth}
Robin Magnet, Jing Ren, Olga Sorkine-Hornung, and Maks Ovsjanikov.
\newblock {Smooth non-rigid shape matching via effective Dirichlet energy optimization}.
\newblock In \emph{International Conference on 3D Vision}, pages 495--504, 2022.

\bibitem[Mahmood et~al.(2019)Mahmood, Ghorbani, Troje, Pons-Moll, and Black]{mahmood2019amass}
Naureen Mahmood, Nima Ghorbani, Nikolaus~F Troje, Gerard Pons-Moll, and Michael~J Black.
\newblock {AMASS: Archive of motion capture as surface shapes}.
\newblock In \emph{IEEE/CVF International Conference on Computer Vision}, pages 5442--5451, 2019.

\bibitem[Markley et~al.(2007)Markley, Cheng, Crassidis, and Oshman]{markley2007averaging}
F~Landis Markley, Yang Cheng, John~L Crassidis, and Yaakov Oshman.
\newblock Averaging quaternions.
\newblock \emph{Journal of Guidance, Control, and Dynamics}, 30\penalty0 (4):\penalty0 1193--1197, 2007.

\bibitem[Martinelli et~al.(2024)Martinelli, Garau, Bisagno, and Conci]{martinelli2024skeleton}
Giulia Martinelli, Nicola Garau, Niccol{\'o} Bisagno, and Nicola Conci.
\newblock Skeleton-aware motion retargeting using masked pose modeling.
\newblock In \emph{European Conference on Computer Vision}, pages 287--303, 2024.

\bibitem[Mohlin et~al.(2020)Mohlin, Sullivan, and Bianchi]{mohlin2020probabilistic}
David Mohlin, Josephine Sullivan, and G{\'e}rald Bianchi.
\newblock Probabilistic orientation estimation with matrix fisher distributions.
\newblock \emph{Advances in Neural Information Processing Systems}, 33:\penalty0 4884--4893, 2020.

\bibitem[Mosella{-}Montoro and Hidalgo(2022)]{mosella2022skinningnet}
Albert Mosella{-}Montoro and Javier~Ruiz Hidalgo.
\newblock {SkinningNet: Two-Stream Graph Convolutional Neural Network for Skinning Prediction of Synthetic Characters}.
\newblock In \emph{IEEE/CVF Conference on Computer Vision and Pattern Recognition}, pages 18593--18602, 2022.

\bibitem[Muralikrishnan et~al.(2025)Muralikrishnan, Dutt, and Mitra]{muralikrishnan2025smf}
Sanjeev Muralikrishnan, Niladri~Shekhar Dutt, and Niloy~J Mitra.
\newblock {SMF: Template-free and Rig-free Animation Transfer using Kinetic Codes}.
\newblock \emph{{ACM} Transactions on Graphics}, 44\penalty0 (6):\penalty0 262:1--262:11, 2025.

\bibitem[Oshman and Carmi(2006)]{oshman2006attitude}
Yaakov Oshman and Avishy Carmi.
\newblock Attitude estimation from vector observations using a genetic-algorithm-embedded quaternion particle filter.
\newblock \emph{Journal of Guidance, Control, and Dynamics}, 29\penalty0 (4):\penalty0 879--891, 2006.

\bibitem[Pai et~al.(2021)Pai, Ren, Melzi, Wonka, and Ovsjanikov]{pai2021fast}
Gautam Pai, Jing Ren, Simone Melzi, Peter Wonka, and Maks Ovsjanikov.
\newblock {Fast Sinkhorn Filters: Using Matrix Scaling for Non-Rigid Shape Correspondence With Functional Maps}.
\newblock In \emph{IEEE Conference on Computer Vision and Pattern Recognition}, pages 384--393, 2021.

\bibitem[Pan et~al.(2021)Pan, Huang, Mai, Wang, Li, Su, Wang, and Jin]{pan2021heterskinnet}
Xiaoyu Pan, Jiancong Huang, Jiaming Mai, He Wang, Honglin Li, Tongkui Su, Wenjun Wang, and Xiaogang Jin.
\newblock {HeterSkinNet: A Heterogeneous Network for Skin Weights Prediction}.
\newblock \emph{Proceedings of the ACM on Computer Graphics and Interactive Techniques}, 4\penalty0 (1):\penalty0 10:1--10:19, 2021.

\bibitem[Paszke et~al.(2019)Paszke, Gross, Massa, Lerer, Bradbury, Chanan, Killeen, Lin, Gimelshein, Antiga, et~al.]{paszke2019pytorch}
Adam Paszke, Sam Gross, Francisco Massa, Adam Lerer, James Bradbury, Gregory Chanan, Trevor Killeen, Zeming Lin, Natalia Gimelshein, Luca Antiga, et~al.
\newblock {PyTorch: An Imperative Style, High-Performance Deep Learning Library}.
\newblock \emph{Advances in Neural Information Processing Systems}, 32:\penalty0 8024--8035, 2019.

\bibitem[Pavlakos et~al.(2019)Pavlakos, Choutas, Ghorbani, Bolkart, Osman, Tzionas, and Black]{pavlakos2019expressive}
Georgios Pavlakos, Vasileios Choutas, Nima Ghorbani, Timo Bolkart, Ahmed~AA Osman, Dimitrios Tzionas, and Michael~J Black.
\newblock {Expressive Body Capture: 3D Hands, Face, and Body From a Single Image}.
\newblock In \emph{IEEE Conference on Computer Vision and Pattern Recognition}, pages 10975--10985, 2019.

\bibitem[Raab et~al.(2024)Raab, Leibovitch, Tevet, Arar, Bermano, and Cohen-Or]{raab2024single}
Sigal Raab, Inbal Leibovitch, Guy Tevet, Moab Arar, Amit~H Bermano, and Daniel Cohen-Or.
\newblock Single motion diffusion.
\newblock In \emph{International Conference on Learning Representations}, 2024.

\bibitem[Radford et~al.(2021)Radford, Kim, Hallacy, Ramesh, Goh, Agarwal, Sastry, Askell, Mishkin, Clark, Krueger, and Sutskever]{radford2021learning}
Alec Radford, Jong~Wook Kim, Chris Hallacy, Aditya Ramesh, Gabriel Goh, Sandhini Agarwal, Girish Sastry, Amanda Askell, Pamela Mishkin, Jack Clark, Gretchen Krueger, and Ilya Sutskever.
\newblock Learning transferable visual models from natural language supervision.
\newblock In \emph{International Conference on Machine Learning}, pages 8748--8763, 2021.

\bibitem[Sengupta et~al.(2021)Sengupta, Budvytis, and Cipolla]{sengupta2021hierarchical}
Akash Sengupta, Ignas Budvytis, and Roberto Cipolla.
\newblock {Hierarchical kinematic probability distributions for 3D human shape and pose estimation from images in the wild}.
\newblock In \emph{IEEE/CVF International Conference on Computer Vision}, pages 11219--11229, 2021.

\bibitem[Sinkhorn and Knopp(1967)]{sinkhorn1967concerning}
Richard Sinkhorn and Paul Knopp.
\newblock Concerning nonnegative matrices and doubly stochastic matrices.
\newblock \emph{Pacific Journal of Mathematics}, 21\penalty0 (2):\penalty0 343--348, 1967.

\bibitem[Song et~al.(2021)Song, Wei, Li, Liu, and Lin]{song20213d}
Chaoyue Song, Jiacheng Wei, Ruibo Li, Fayao Liu, and Guosheng Lin.
\newblock {3D} pose transfer with correspondence learning and mesh refinement.
\newblock \emph{Advances in Neural Information Processing Systems}, 34:\penalty0 3108--3120, 2021.

\bibitem[Song et~al.(2025{\natexlab{a}})Song, Li, Yang, Xu, Wei, Liu, Feng, Lin, and Zhang]{song2025puppeteer}
Chaoyue Song, Xiu Li, Fan Yang, Zhongcong Xu, Jiacheng Wei, Fayao Liu, Jiashi Feng, Guosheng Lin, and Jianfeng Zhang.
\newblock {Puppeteer: Rig and Animate Your 3D Models}.
\newblock \emph{Advances in neural information processing systems}, 2025{\natexlab{a}}.

\bibitem[Song et~al.(2025{\natexlab{b}})Song, Zhang, Li, Yang, Chen, Xu, Liew, Guo, Liu, Feng, and Guosheng]{song2025magicarticulate}
Chaoyue Song, Jianfeng Zhang, Xiu Li, Fan Yang, Yiwen Chen, Zhongcong Xu, Jun~Hao Liew, Xiaoyang Guo, Fayao Liu, Jiashi Feng, and Lin Guosheng.
\newblock {MagicArticulate: Make Your 3D Models Articulation-Ready}.
\newblock In \emph{IEEE/CVF Conference on Computer Vision and Pattern Recognition}, pages 15998--16007, 2025{\natexlab{b}}.

\bibitem[Sorkine and Alexa(2007)]{sorkine2007rigid}
Olga Sorkine and Marc Alexa.
\newblock As-rigid-as-possible surface modeling.
\newblock In \emph{Symposium on Geometry processing}, pages 109--116, 2007.

\bibitem[Sumner and Popovi{\'c}(2004)]{sumner2004deformation}
Robert~W Sumner and Jovan Popovi{\'c}.
\newblock Deformation transfer for triangle meshes.
\newblock \emph{ACM Transactions on Graphics}, 23\penalty0 (3):\penalty0 399--405, 2004.

\bibitem[Tagliasacchi et~al.(2009)Tagliasacchi, Zhang, and Cohen{-}Or]{tagliasacchi2009curve}
Andrea Tagliasacchi, Hao Zhang, and Daniel Cohen{-}Or.
\newblock Curve skeleton extraction from incomplete point cloud.
\newblock \emph{ACM Transactions on Graphics}, 28\penalty0 (3):\penalty0 71:1--71:9, 2009.

\bibitem[{Truebones Motions Animation Studios}(2022)]{truebones2022}
{Truebones Motions Animation Studios}.
\newblock Truebones motion capture library.
\newblock Online service by Truebones Motions Animation Studios, 2022.
\newblock Accessed January 2025.

\bibitem[Vaswani et~al.(2017)Vaswani, Shazeer, Parmar, Uszkoreit, Jones, Gomez, Kaiser, and Polosukhin]{vaswani2017attention}
Ashish Vaswani, Noam Shazeer, Niki Parmar, Jakob Uszkoreit, Llion Jones, Aidan~N Gomez, {\L}ukasz Kaiser, and Illia Polosukhin.
\newblock Attention is all you need.
\newblock \emph{Advances in Neural Information Processing Systems}, 30:\penalty0 5998--6008, 2017.

\bibitem[Wang et~al.(2020)Wang, Wen, Fu, Lin, Zou, Xue, and Zhang]{wang2020neural}
Jiashun Wang, Chao Wen, Yanwei Fu, Haitao Lin, Tianyun Zou, Xiangyang Xue, and Yinda Zhang.
\newblock Neural pose transfer by spatially adaptive instance normalization.
\newblock In \emph{IEEE/CVF Conference on Computer Vision and Pattern Recognition}, pages 5831--5839, 2020.

\bibitem[Wang et~al.(2023{\natexlab{a}})Wang, Li, Liu, De~Mello, Gallo, Wang, and Kautz]{wang2023zero}
Jiashun Wang, Xueting Li, Sifei Liu, Shalini De~Mello, Orazio Gallo, Xiaolong Wang, and Jan Kautz.
\newblock {Zero-shot pose transfer for unrigged stylized 3D characters}.
\newblock In \emph{Proceedings of the IEEE/CVF Conference on Computer Vision and Pattern Recognition}, pages 8704--8714, 2023{\natexlab{a}}.

\bibitem[Wang et~al.(2019)Wang, Yan, and Yang]{wang2019learning}
Runzhong Wang, Junchi Yan, and Xiaokang Yang.
\newblock Learning combinatorial embedding networks for deep graph matching.
\newblock In \emph{IEEE/CVF International Conference on Computer Vision}, pages 3056--3065, 2019.

\bibitem[Wang et~al.(2023{\natexlab{b}})Wang, Yan, and Yang]{wang2020combinatorial}
Runzhong Wang, Junchi Yan, and Xiaokang Yang.
\newblock {Combinatorial Learning of Robust Deep Graph Matching: An Embedding Based Approach}.
\newblock \emph{IEEE Transactions on Pattern Analysis and Machine Intelligence}, 45\penalty0 (6):\penalty0 6984--7000, 2023{\natexlab{b}}.

\bibitem[Wang et~al.(2024)Wang, Mao, Lu, and Li]{wang2024towards}
Rong Wang, Wei Mao, Changsheng Lu, and Hongdong Li.
\newblock {Towards high-quality 3D motion transfer with realistic apparel animation}.
\newblock In \emph{European Conference on Computer Vision}, pages 35--51, 2024.

\bibitem[Wu et~al.(2022)Wu, Chen, Liu, Ren, and Wang]{wu2022casa}
Yuefan Wu, Zeyuan Chen, Shaowei Liu, Zhongzheng Ren, and Shenlong Wang.
\newblock {CASA: Category-agnostic Skeletal Animal Reconstruction}.
\newblock \emph{Advances in Neural Information Processing Systems}, 35:\penalty0 28559--28574, 2022.

\bibitem[Xu et~al.(2022)Xu, Li, Yang, Shi, Fu, and Huang]{xu2022hierarchical}
Pengfei Xu, Yifan Li, Zhijin Yang, Weiran Shi, Hongbo Fu, and Hui Huang.
\newblock Hierarchical layout blending with recursive optimal correspondence.
\newblock \emph{ACM Transactions on Graphics}, 41\penalty0 (6):\penalty0 249:1--249:15, 2022.

\bibitem[Xu et~al.(2019)Xu, Zhou, Kalogerakis, and Singh]{xu2019predicting}
Zhan Xu, Yang Zhou, Evangelos Kalogerakis, and Karan Singh.
\newblock {Predicting Animation Skeletons for 3D Articulated Models via Volumetric Nets}.
\newblock In \emph{International Conference on 3D Vision}, pages 298--307, 2019.

\bibitem[Xu et~al.(2020)Xu, Zhou, Kalogerakis, Landreth, and Singh]{xu2020rignet}
Zhan Xu, Yang Zhou, Evangelos Kalogerakis, Chris Landreth, and Karan Singh.
\newblock {RigNet: neural rigging for articulated characters}.
\newblock \emph{ACM Transactions on Graphics}, 39\penalty0 (4):\penalty0 58:1--58:14, 2020.

\bibitem[Yifan et~al.(2020)Yifan, Aigerman, Kim, Chaudhuri, and Sorkine{-}Hornung]{yifan2020neural}
Wang Yifan, Noam Aigerman, Vladimir~G Kim, Siddhartha Chaudhuri, and Olga Sorkine{-}Hornung.
\newblock {Neural Cages for Detail-Preserving 3D Deformations}.
\newblock In \emph{IEEE/CVF Conference on Computer Vision and Pattern Recognition}, pages 75--83, 2020.

\bibitem[Yin et~al.(2022)Yin, Cai, Wang, and Chen]{yin2022fishermatch}
Yingda Yin, Yingcheng Cai, He Wang, and Baoquan Chen.
\newblock {FisherMatch: Semi-Supervised Rotation Regression via Entropy-based Filtering}.
\newblock In \emph{IEEE/CVF Conference on Computer Vision and Pattern Recognition}, pages 11164--11173, 2022.

\bibitem[Yoo et~al.(2024)Yoo, Koo, Yeo, and Sung]{yoo2024neural}
Seungwoo Yoo, Juil Koo, Kyeongmin Yeo, and Minhyuk Sung.
\newblock Neural pose representation learning for generating and transferring non-rigid object poses.
\newblock \emph{Advances in Neural Information Processing Systems}, 38:\penalty0 34349--34377, 2024.

\bibitem[Yu et~al.(2025)Yu, Wang, Wang, Zhang, Liu, Li, Ni, and Zhang]{yu2025mesh2animation}
Zhenbo Yu, Junjie Wang, Hang Wang, Zhiyuan Zhang, Jinxian Liu, Zefan Li, Bingbing Ni, and Wenjun Zhang.
\newblock {Mesh2Animation: Unsupervised Animating for Quadruped 3D Objects}.
\newblock \emph{IEEE Transactions on Circuits and Systems for Video Technology}, 35\penalty0 (6):\penalty0 5711--5723, 2025.

\bibitem[Yun et~al.(2025)Yun, Hong, Kim, and Noh]{yun2025anymole}
Kwan Yun, Seokhyeon Hong, Chaelin Kim, and Junyong Noh.
\newblock {AnyMoLe: Any Character Motion In-betweening Leveraging Video Diffusion Models}.
\newblock In \emph{IEEE/CVF Conference on Computer Vision and Pattern Recognition}, pages 27838--27848, 2025.

\bibitem[Zhan et~al.(2024)Zhan, Fu, and Ritchie]{zhan2024charactermixer}
Xiao Zhan, Rao Fu, and Daniel Ritchie.
\newblock {CharacterMixer: Rig-Aware Interpolation of 3D Characters}.
\newblock \emph{Computer Graphics Forum}, 43\penalty0 (2), 2024.

\bibitem[Zhang et~al.(2025{\natexlab{a}})Zhang, Chang, Li, Soleymani, and Ahuja]{zhangmagicpose4d}
Hao Zhang, Di Chang, Fang Li, Mohammad Soleymani, and Narendra Ahuja.
\newblock {MagicPose4D: Crafting Articulated Models with Appearance and Motion Control}.
\newblock \emph{Transactions on Machine Learning Research}, 2025{\natexlab{a}}.

\bibitem[Zhang et~al.(2025{\natexlab{b}})Zhang, Xu, Feng, Jampani, and Ahuja]{zhang2025physrig}
Hao Zhang, Haolan Xu, Chun Feng, Varun Jampani, and Narendra Ahuja.
\newblock {PhysRig: Differentiable Physics-Based Skinning and Rigging Framework for Realistic Articulated Object Modeling}.
\newblock In \emph{IEEE/CVF International Conference on Computer Vision}, pages 6609--6620, 2025{\natexlab{b}}.

\bibitem[Zhang et~al.(2024)Zhang, Huang, Tu, Chen, Zhan, YU, and Shan]{zhang2024tapmo}
Jiaxu Zhang, Shaoli Huang, Zhigang Tu, Xin Chen, Xiaohang Zhan, Gang YU, and Ying Shan.
\newblock {TapMo: Shape-aware Motion Generation of Skeleton-free Characters}.
\newblock In \emph{International Conference on Learning Representations}, 2024.

\bibitem[Zhang et~al.(2025{\natexlab{c}})Zhang, Pu, Guo, Cao, and Hu]{zhang2025one}
Jia-Peng Zhang, Cheng-Feng Pu, Meng-Hao Guo, Yan-Pei Cao, and Shi-Min Hu.
\newblock One model to rig them all: Diverse skeleton rigging with unirig.
\newblock \emph{ACM Transactions on Graphics}, 44\penalty0 (4):\penalty0 123:1--123:18, 2025{\natexlab{c}}.

\bibitem[Zhang et~al.(2020)Zhang, Zhang, Chen, Yuan, and Wen]{zhang2020cross}
Pan Zhang, Bo Zhang, Dong Chen, Lu Yuan, and Fang Wen.
\newblock Cross-domain correspondence learning for exemplar-based image translation.
\newblock In \emph{IEEE/CVF Conference on Computer Vision and Pattern Recognition}, pages 5143--5153, 2020.

\bibitem[Zhao et~al.(2024)Zhao, Li, Yifan, Olga, and Wetzstein]{zhao2024pose}
Qingqing Zhao, Peizhuo Li, Wang Yifan, Sorkine-Hornung Olga, and Gordon Wetzstein.
\newblock Pose-to-motion: Cross-domain motion retargeting with pose prior.
\newblock In \emph{Computer Graphics Forum}, page e15170. Wiley Online Library, 2024.

\bibitem[Zhao et~al.(2023)Zhao, Liu, Chen, Zeng, Wang, Cheng, Fu, Chen, Yu, and Gao]{zhao2023michelangelo}
Zibo Zhao, Wen Liu, Xin Chen, Xianfang Zeng, Rui Wang, Pei Cheng, Bin Fu, Tao Chen, Gang Yu, and Shenghua Gao.
\newblock {Michelangelo: Conditional 3D Shape Generation based on Shape-Image-Text Aligned Latent Representation}.
\newblock \emph{Advances in Neural Information Processing Systems}, 36:\penalty0 73969--73982, 2023.

\bibitem[Zhou et~al.(2020)Zhou, Bhatnagar, and Pons-Moll]{zhou2020unsupervised}
Keyang Zhou, Bharat~Lal Bhatnagar, and Gerard Pons-Moll.
\newblock {Unsupervised Shape and Pose Disentanglement for 3D Meshes}.
\newblock In \emph{European Conference on Computer Vision}, pages 341--357. Springer, 2020.

\end{thebibliography}
